%% file: main.tex
\documentclass[10pt,letterpaper]{article}
\usepackage{neutralpreprint}
\usepackage{amsmath,amssymb,booktabs,graphicx}
\usepackage[numbers,sort&compress]{natbib}
\usepackage{caption,needspace}
\usepackage[hyphens]{url}
\usepackage[hidelinks]{hyperref}
\newcommand{\method}{Semantic Prism}
\newcommand{\hgea}{HGEA}
\newcommand{\cihd}{C-IHD}
\newcommand{\R}{\mathbb{R}}
\newcommand{\PaperTitle}{Semantic Prism: Generative Segmentation with an Explicit Semantic-Image Interface}
\input{author_info}

\title{\PaperTitle}
\author{\PaperAuthors}
\affiliations{\PaperAffiliations}
\date{}
\hypersetup{pdftitle={\PaperTitle},pdfauthor={\PaperPdfAuthors}}

\newcommand{\makesupplementtitle}{%
  \clearpage
  \twocolumn[{%
    \centering
    {\LARGE\bfseries Supplementary Material\par}
    \vspace{5pt}
    {\large\bfseries \PaperTitle\par}
    \vspace{8pt}
    {\large\bfseries \PaperAuthors\par}
    \vspace{5pt}
    {\normalsize \PaperAffiliations\par}
    \vspace{15pt}
  }]%
  \thispagestyle{plain}%
}

\begin{document}
\maketitle
\thispagestyle{plain}
\input{main_body}
\FloatBarrier
\Needspace{9\baselineskip}
\input{llm_usage_statement}
{\small
\bibliographystyle{ieeenat_fullname}
\bibliography{references}
}

\clearpage
\appendix
\setcounter{section}{0}
\setcounter{subsection}{0}
\setcounter{figure}{0}
\setcounter{table}{0}
\setcounter{equation}{0}
\renewcommand{\thesection}{S\arabic{section}}
\renewcommand{\thesubsection}{\thesection.\arabic{subsection}}
\renewcommand{\thefigure}{S\arabic{figure}}
\renewcommand{\thetable}{S\arabic{table}}
\renewcommand{\theequation}{S\arabic{equation}}
\renewcommand{\theHsection}{supp.\arabic{section}}
\renewcommand{\theHsubsection}{supp.\arabic{section}.\arabic{subsection}}
\renewcommand{\theHfigure}{supp.\arabic{figure}}
\renewcommand{\theHtable}{supp.\arabic{table}}
\renewcommand{\theHequation}{supp.\arabic{equation}}
\makesupplementtitle
\input{supporting_materials}
\FloatBarrier
\end{document}

%% file: author_info.tex
\newcommand{\PaperAuthors}{%
  Weize Cai$^{1,\dagger}$ \quad
  Yongqi Dong$^{1,2,\dagger,*}$ \quad
  Zhida Shao$^{1,\dagger}$ \quad
  Zixin Fu$^{3,\dagger}$ \quad
  Xiaoxi Hu$^{4}$%
}
\newcommand{\PaperPdfAuthors}{Weize Cai, Yongqi Dong, Zhida Shao, Zixin Fu, Xiaoxi Hu}
\newcommand{\PaperAffiliations}{%
  $^{1}$RWTH Aachen University, Aachen, Germany\\
  $^{2}$Delft University of Technology, Delft, The Netherlands\\
  $^{3}$Chang'an University, Xi'an, China\\
  $^{4}$State Key Laboratory of Intelligent Green Vehicle and Mobility,\\
  Tsinghua University, China\\[3pt]
  $^{\dagger}$Equal contribution.\quad
  $^{*}$Corresponding author: \texttt{yongqi.dong@rwth-aachen.de}%
}

%% file: main_body.tex
\begin{abstract}
Conditional image generators can render segmentation maps as semantic RGB images, but color drift and boundary mixing can undermine direct decoding. We present Semantic Prism, a framework that combines an explicit semantic-image interface with hierarchical refinement for road-scene segmentation. A generator adapted from a diffusion-distilled backbone produces a semantic RGB image with a single denoising update per crop. Distances to a fixed class-color codebook define interface logits and pixelwise class probabilities. Hierarchical Generator Evidence Alignment (HGEA) combines spatially aligned multilevel features from the VAE encoder of the generator with the input RGB image, generated semantic image, and interface probabilities to predict additive corrections to these logits. Contextual Interface Hierarchy Disagreement (C-IHD) combines pointwise uncertainty, local context, and interface-refinement disagreement in a fixed score for ranking pixel errors, without an auxiliary predictor or an additional network pass. On the 500-image Cityscapes validation set, Semantic Prism achieves 72.07\% mean intersection over union (mIoU), improving on direct-interface decoding by 11.39 percentage points. Extensive capacity-matched ablations show higher mean mIoU for multilevel refinement than for flat and single-level alternatives. The Cityscapes-trained model transfers to the Adverse Conditions Dataset with Correspondences (ACDC), achieving 46.89\% mIoU without target-domain adaptation. Beyond Cityscapes, on the BDD100K dataset, a separately trained model achieves 62.22\% mIoU on its full 1,000-image validation set. Across the three evaluated datasets, C-IHD improves pixel-error area under the precision-recall curve (AUPR) over uncertainty derived from maximum softmax probability, with segmentation predictions held fixed for each comparison. In particular, on ACDC, AUPR increases from 0.6580 to 0.7557, demonstrating that Semantic Prism presents more effective error ranking under adverse-condition domain shift.

\end{abstract}

\section{Introduction}

Semantic segmentation supports scene understanding by assigning a semantic class to every image pixel. Modern discriminative architectures achieve strong accuracy by mapping latent image representations directly to class logits \citep{xie2021segformer,cheng2022mask2former}. Generative formulations cast dense prediction as the generation of a label map or semantic image \citep{chen2023gss,lai2023ddps,ji2023ddp}, exposing a visible structured output from which pixelwise class labels can be decoded. However, combining an explicitly decodable semantic-image interface that can be evaluated independently of latent generator features with fine-grained spatial refinement remains challenging.

Direct color decoding provides an explicit mapping between rendered colors and predicted classes. However, color deviations, mixed colors near class boundaries, and blur can lead to incorrect class assignments, particularly around semantic boundaries and in thin structures. Latent-feature decoders can recover local detail from generator representations. This motivates a formulation in which the semantic image defines a per-pixel class distribution, while aligned multi-level generator features provide additive corrections to the corresponding class logits.

We realize this design with \method{}, a one-step semantic-image generation-and-refinement framework with deterministic inference. A pix2pix-Turbo image translator \citep{parmar2024onestep,sauer2023add} renders a semantic RGB image from the input image. At each pixel, distances from the rendered RGB value to a fixed class-color codebook define a full class-probability distribution, forming an explicit probabilistic interface. We call this interface \emph{observable} in an operational sense: the pre-refinement distribution, including its top-1 label, maximum class probability, and pairwise class log-odds, is recoverable from the rendered image using the fixed codebook decoder, without access to latent generator features.

Hierarchical Generator Evidence Alignment (\hgea{}) complements this interface by spatially aligning features from three generator levels. A zero-initialized output projection is used to predict additive corrections to the interface logits. These corrections use the input image, rendered semantic image, interface distribution, and aligned generator features. As an auxiliary readout, Contextual Interface Hierarchy Disagreement (C-IHD) combines pointwise and local uncertainty with disagreement between the interface and refined distributions to rank remaining pixel errors. It leaves the segmentation unchanged and requires neither a trainable error predictor nor an additional model forward pass.

We evaluate \method{} in three complementary settings: in-domain evaluation on Cityscapes \citep{cordts2016cityscapes}, independent in-domain training and evaluation on BDD100K \citep{yu2020bdd100k}, and source-frozen transfer from Cityscapes to the Adverse Conditions Dataset with Correspondences (ACDC) \citep{sakaridis2021acdc}. These experiments assess segmentation accuracy, boundary quality, calibration, and pixel-error ranking.

Our contributions are threefold: (1) We formulate an observable semantic-image interface whose fixed distance-based codebook decoder maps rendered RGB values to full class distributions, with top-1 labels directly recoverable from the rendered image and pairwise class log-odds available in closed form; (2) We propose \hgea{}, which aligns multi-level generator features and additively refines interface logits. On Cityscapes, the full refinement module improves mean intersection over union (mIoU) by 11.39 percentage points over direct-interface decoding, reaching 72.07\%; matched-capacity three-seed controls show higher mIoU for multi-level refinement than for flat and single-level refiners; (3) We introduce \cihd{}, a fixed readout that combines pointwise uncertainty, local context, and interface--refinement disagreement to improve pixel-error ranking, measured by the area under the precision--recall curve (AUPR), relative to maximum softmax probability (MSP) on the same underlying predictions in each of the three evaluation settings.

\section{Related work}

\paragraph{Discriminative semantic segmentation.}
Modern segmentation-based methods combine local detail, multiscale context, and global interaction through convolutional decoders \citep{chen2018deeplabv3plus}, hierarchical transformers \citep{xie2021segformer}, mask classification \citep{cheng2022mask2former}, and state-space models \citep{fu2025segman}. Domain-generalized methods target unseen appearance shifts through feature regularization, generative guidance, or generated training data \citep{choi2021robustnet,li2025querydiff,jia2024dginstyle}. These methods produce final predictions with learned heads operating on latent features. We study a complementary output construction in which the rendered semantic image, together with a fixed decoder, defines a pre-refinement probability field that can be evaluated without access to latent features.

\paragraph{Generative and image-form dense prediction.}
An alternative line of work represents dense predictions directly in image space. GSS encodes segmentation masks as RGB ``maskige'' images \citep{chen2023gss}, Painter represents visual-task outputs as images \citep{wang2023painter}, SegGPT performs in-context coloring \citep{wang2023seggpt}, and UniGS \citep{qi2024unigs} and CAM-Seg \citep{ahmed2025camseg} introduce location-aware or continuous semantic-image representations. SemFlow models segmentation and image synthesis as reverse transports between image and mask distributions \citep{wang2024semflow}. Other generative approaches employ diffusion, flow, or autoregressive processes to generate or refine dense label maps \citep{ji2023ddp,lai2023ddps,wang2023segrefiner,caetano2026symmflow,yang2026genmask,deng2025llamaseg}, while generalist models formulate diverse perception tasks as conditional image generation \citep{geng2024instructdiffusion,zhao2025diception}. Vision Banana represents semantic segmentation as an RGB image using prompt-specified class colors \citep{gabeur2026visionbanana}. Collectively, these works demonstrate image-form dense prediction and, in some cases, color-based label recovery. \method{} uses fixed codebook distances to define a full per-pixel class distribution from the rendered RGB image and parameterizes refinement as additive corrections in the resulting interface logit space.

\paragraph{Generator representations and inspectable interfaces.}
Beyond image-form outputs, prior work has explored internal generative representations and structured intermediate interfaces. VPD \citep{zhao2023vpd} and ODISE \citep{xu2023odise} couple diffusion features with learned segmentation decoders, while training-free methods \citep{meng2026trainingfree} aggregate internal features and attention maps. Concept and semantic bottlenecks expose semantically aligned intermediate variables \citep{koh2020conceptbottleneck,losch2021semanticbottlenecks}, whereas prototype segmentors relate predictions to learned examples or parts \citep{sacha2023protoseg,porta2025scaleprotoseg}. In our formulation, a fixed class-color decoder maps the rendered semantic image to an independently evaluable pre-refinement distribution. \hgea{} uses aligned generator features together with image and interface inputs to additively correct the interface logits.

\paragraph{Predictive reliability and error ranking.}
In addition to segmentation accuracy, predictive reliability concerns both calibration
and the ranking of prediction errors. Expected calibration error (ECE)
summarizes the discrepancy between predictive confidence and empirical accuracy
\citep{guo2017calibration}, whereas the complement of maximum softmax
probability (MSP) provides a standard deterministic score for ranking errors
\citep{hendrycks2017baseline}. Dense calibration methods learn pixelwise or
multivariate confidence corrections
\citep{ding2021localtemperature,wang2023calibrating,kuppers2022confidence},
while neighbor-aware objectives incorporate local spatial structure during
training \citep{murugesan2025neighbor}. Failure-localization networks introduce
auxiliary predictors \citep{rahman2022fsnet,kwon2022eln}, whereas segment- and
image-level methods aggregate uncertainty for quality estimation
\citep{rottmann2020metaclassification,guarino2026spatialaggregation}. Selective
prediction evaluates such rankings through the risk--coverage trade-off
\citep{geifman2017selective}. The proposed \cihd{} instead targets pixel-error ranking on fixed predictions, requiring
neither an auxiliary predictor nor an additional forward pass.

\section{Method}

\subsection{Overview}
Given an input image $\mathbf{x}\in\R^{H\times W\times3}$ on the pixel lattice
$\Omega=\{1,\ldots,H\}\times\{1,\ldots,W\}$, a one-step conditional generator
deterministically renders a semantic RGB image $\mathbf{s}$ and exposes
intermediate feature maps
$\mathcal H_{\mathbf{x}}=\{\mathbf h^\ell_{\mathbf{x}}\}_{\ell\in\mathcal J}$,
where $\mathcal J$ indexes the selected generator levels and
$\mathbf h^\ell_{\mathbf{x}}$ denotes the feature map extracted at level
$\ell$.

Let $\mathcal C=\{\mathbf c_k\}_{k=1}^{K}$ be a fixed
high-separation class-color codebook, where $K$ is the number of semantic
classes and $\mathbf c_k\in\mathbb R^3$ is the RGB prototype assigned per class
$k$. \method{} forms two coupled distributions. A fixed distance-based decoder maps
the rendered semantic image to an interface distribution, while \hgea{} uses
the feature hierarchy to predict an additive residual in the same logit space.

For each pixel $u\in\Omega$, the fixed decoder first maps the rendered semantic image to an interface distribution. \hgea{} then applies a learned residual to the corresponding class logits:
\begin{align}
p_I(u)
&=\Pi_{\mathcal C}^{(\tau_I)}(\mathbf s(u))
=\operatorname{softmax}(\mathbf z^I(u)),
\label{eq:overview-interface}\\
p_H(u)
&=\operatorname{softmax}\!\left(
\mathbf z^I(u)+\Delta\mathbf z^H(u)
\right).
\label{eq:overview-refinement}
\end{align}
Here $\Pi_{\mathcal C}^{(\tau_I)}$ denotes the fixed distance-based codebook decoder, $\mathbf z^I$ denotes the prototype-distance logits underlying $p_I$, and $\Delta\mathbf z^H$ denotes the learned refinement residual. The rendered image and fixed decoder determine the interface distribution and its pairwise class log-odds; \hgea{} predicts additive corrections to the same logits. For each pixel $u\in\Omega$, the final label is $\hat y(u)=\arg\max_k p_{H,k}(u)$. MSP and \cihd{} are computed downstream of this prediction path and leave both refined distribution $p_H$ and the predicted label map $\hat y$ unchanged.

\begin{figure*}[t]
\centering
\includegraphics[width=\textwidth]{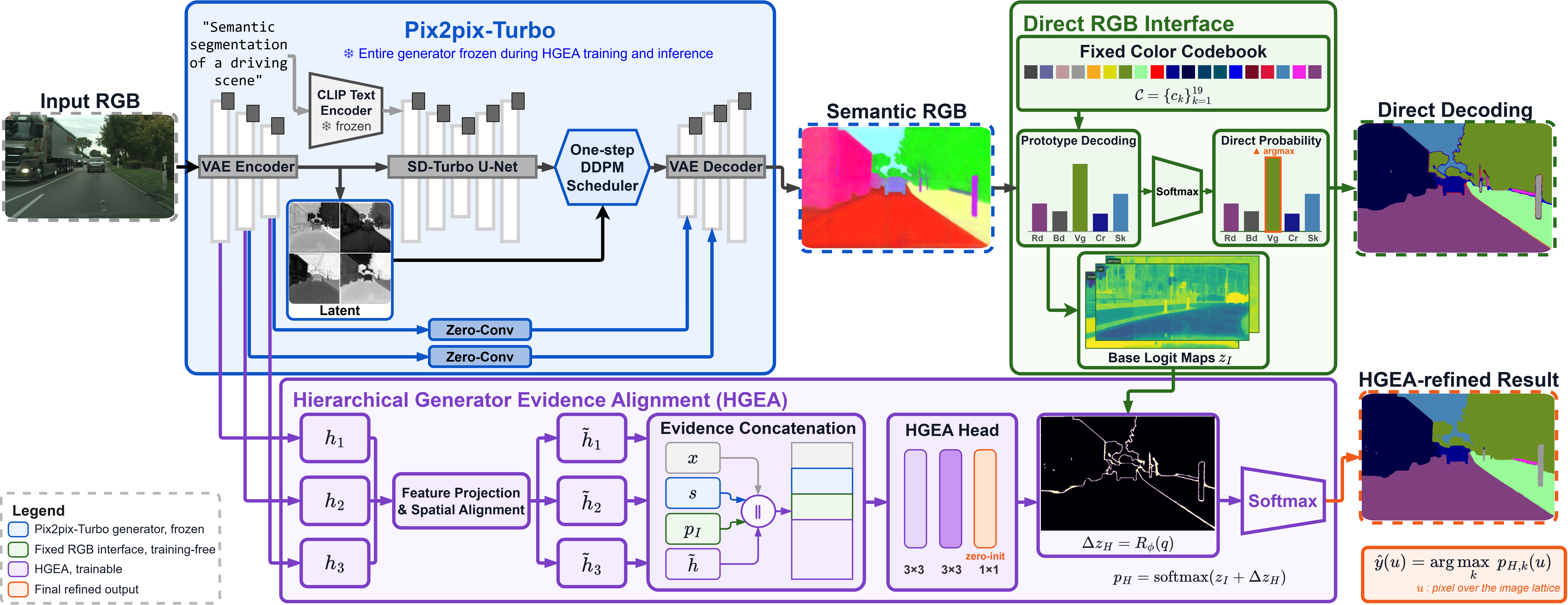}
\caption{Overview of \method{}. A one-step pix2pix-Turbo generator renders a
semantic RGB image and exposes multi-level internal features. The fixed
codebook decodes the image into the interface distribution $p_I$, while
\hgea{} aligns the feature hierarchy and predicts
$\Delta\mathbf z^H$, yielding the refined distribution $p_H$.}
\label{fig:framework1}
\end{figure*}

Figure~\ref{fig:framework1} summarizes the complete pipeline.

\subsection{One-step semantic-image generation}

We encode a label field $\mathbf y$ as a semantic target image $\mathbf s^\star(u)=\mathbf c_{y(u)}$ and adapt the pix2pix-Turbo generator \citep{parmar2024onestep,sauer2023add} using a fixed segmentation prompt. SD-Turbo provides its pretrained backbone, and inference uses one generator step per crop. The codebook is constructed once by greedy max--min selection on a saturated RGB grid, followed by confusion-aware class assignment; the exact prototypes are listed in Table~\ref{tab:supp-codebook}. During generator training, rendered colors and prototypes are expressed in normalized RGB coordinates. For a color $\mathbf v$ and temperature $\tau>0$, the fixed prototype decoder is
\begin{equation}
\Pi_{\mathcal C}^{(\tau)}(\mathbf v)_k
=
\frac{\exp\!\left(-\|\mathbf v-\mathbf c_k\|_2^2/\tau\right)}
{\sum_j \exp\!\left(-\|\mathbf v-\mathbf c_j\|_2^2/\tau\right)}.
\label{eq:prototype-energy}
\end{equation}

The generator objective is
\begin{equation}
\begin{aligned}
\mathcal L_G={}&
0.5\mathcal L_{\mathrm{rgb}}
+3.0\mathcal L_{\mathrm{proto}}
+\mathcal L_{\mathrm{valid}}\\
&+1.5\mathcal L_{\mathrm{margin}}
+0.3\mathcal L_{\mathrm{boundary}}.
\end{aligned}
\label{eq:generator-loss}
\end{equation}

Here $\mathcal L_{\mathrm{rgb}}$ is a masked Smooth-L1 loss between the
generated and target prototype images. For a generated color $\mathbf v$, let
$d_k^{\mathrm G}(\mathbf v)=\|\mathbf v-\mathbf c_k\|_2^2$ denote its squared
distance to prototype $\mathbf c_k$ in normalized RGB space.
$\mathcal L_{\mathrm{proto}}$ is class-weighted cross-entropy over logits
$-d_k^{\mathrm G}(\mathbf v)/\tau_g$, and
$\mathcal L_{\mathrm{valid}}$ averages $\min_k d_k^{\mathrm G}(\mathbf v)$.
The margin loss averages
$[m+d_y^{\mathrm G}(\mathbf v)-\min_{j\ne y}d_j^{\mathrm G}(\mathbf v)]_+$,
encouraging the target prototype to be closer than the nearest competitor by at
least $m$. $\mathcal L_{\mathrm{boundary}}$ applies the same loss within a
radius-2 ground-truth boundary band. Ignore pixels are excluded from all
reductions. Generator supervision uses RGB coordinates normalized to $[0,1]$; thus,
we set the prototype-softmax temperature to $\tau_g=0.03$ and the margin to
$m=0.02$ in normalized squared-distance units.
The loss weights were fixed during
source-only development and were not tuned on validation data.

\subsection{Observable semantic-image interface}

For interface decoding, rendered colors and codebook prototypes are represented
on the $[0,255]$ RGB scale. Because the decoder operates on squared Euclidean
color distances, we set the interface temperature to $\tau_I=900$ in the
corresponding squared-distance units. The fixed
decoder maps each rendered pixel to a full class distribution using Equation (\ref{eq:overview-interface}).
For each $u\in\Omega$, let
$d_k^{\mathrm I}(u)=\|\mathbf s(u)-\mathbf c_k\|_2^2$ and
$z_k^I(u)=-d_k^{\mathrm I}(u)/\tau_I$, so that
$p_I(u)=\operatorname{softmax}(\mathbf z^I(u))$. The rendered RGB value
determines its top-1 label:
\begin{equation}
\hat y_I(u)
=\arg\max_k p_{I,k}(u)
=\arg\min_k d_k^{\mathrm I}(u).
\label{eq:interface-top1}
\end{equation}
For any two distinct classes $a,b\in\{1,\ldots,K\}$, the corresponding
pairwise log-odds are
\begin{equation}
\Lambda^I_{ab}(u)
=\log\frac{p_{I,a}(u)}{p_{I,b}(u)}
=\frac{d_b^{\mathrm I}(u)-d_a^{\mathrm I}(u)}{\tau_I}.
\label{eq:interface-logodds}
\end{equation}

Thus, given the fixed codebook and temperature, the complete interface
distribution is recoverable from the rendered image without access to latent
features. In RGB space, each class's top-1 decision region is its
codebook-induced Euclidean Voronoi cell (i.e., the set of colors nearest to that
class prototype), while the pairwise log-odds provide closed-form relative
evidence between competing classes.

\subsection{Hierarchical Generator Evidence Alignment}

The rendered semantic image may contain blurred class boundaries or omit thin
structures. \hgea{} supplements the interface with spatially aligned multi-level
features from the frozen generator. Let $\mathcal J$ index the selected feature
levels ($|\mathcal J|=3$ in our implementation). For each
$\ell\in\mathcal J$, a $1\times1$ projection
$\mathbf W^\ell_{1\times1}$ maps $\mathbf h^\ell_{\mathbf x}$ to 24 channels,
followed by group normalization (GN), SiLU activation, and bilinear resampling ($\mathcal U_{\ell\rightarrow\Omega}$) to the
output lattice $\Omega$:
\begin{equation}
\widetilde{\mathbf h}^{\ell}
=
\mathcal U_{\ell\rightarrow\Omega}\!\left(
\operatorname{SiLU}\!\left(
\operatorname{GN}\!\left(
\mathbf W^\ell_{1\times1} * \mathbf h^\ell_{\mathbf x}
\right)\right)\right),
\qquad \ell\in\mathcal J.
\label{eq:hgea-align}
\end{equation}

The aligned feature maps are concatenated and combined with the input image,
rendered semantic image, and interface distribution to predict the logit
residual:
\begin{subequations}
\label{eq:hgea-residual}
\begin{align}
\widetilde{\mathbf h}
&=
\operatorname*{Concat}_{\ell\in\mathcal J}
\widetilde{\mathbf h}^{\ell},
\label{eq:hgea-hierarchy}\\
\Delta\mathbf z^H
&=
\mathcal R_\phi\!\left(
\operatorname{Concat}
(\mathbf x,\mathbf s,p_I,\widetilde{\mathbf h})
\right).
\label{eq:hgea-prediction}
\end{align}
\end{subequations}
The residual head $\mathcal R_\phi$ comprises two
$3\times3$ Conv--GN--SiLU blocks followed by a zero-initialized
$1\times1$ output projection that produces $K$ class-logit residuals at each
pixel.

For any two distinct classes $a\ne b$, define
$\Lambda^H_{ab}(u)=\log[p_{H,a}(u)/p_{H,b}(u)]$. The update in
Equation~\eqref{eq:overview-refinement} then yields
\begin{equation}
\Lambda^H_{ab}(u)-\Lambda^I_{ab}(u)
=
\Delta z^H_a(u)-\Delta z^H_b(u).
\label{eq:hgea-logodds}
\end{equation}
Thus, \hgea{} modifies the pairwise interface log-odds through additive class-logit residuals. The zero-initialized output projection ensures $\Delta\mathbf z^H(u)=\mathbf 0$ for every $u\in\Omega$, and hence $p_H(u)=p_I(u)$ at initialization.

With the generator frozen, we train only the 190,891-parameter \hgea{} module
using semantic supervision, boundary- and interior-focused objectives,
target-aware sampling, mild class reweighting, and staged false-positive
suppression. Complete objectives and training schedules are provided in the
supplementary material.

\subsection{Contextual Interface Hierarchy Disagreement}

The complement of the maximum softmax probability (MSP) of the refined
distribution provides a pointwise uncertainty score. \cihd{} combines this
score with local context and interface--refinement disagreement:
\begin{subequations}
\label{eq:cihd-context}
\begin{align}
\mathcal U_{\mathrm{MSP}}(u)
&=
1-\max_k p_{H,k}(u),
\label{eq:cihd-msp}\\
\mathcal U_{\mathrm{loc}}(u)
&=
\mathcal A_{5\times5}
\!\left[\mathcal U_{\mathrm{MSP}}\right](u),
\label{eq:cihd-local}
\end{align}
\end{subequations}
where $\mathcal A_{5\times5}$ denotes a $5\times5$ local averaging operator.

To quantify the distributional change induced by \hgea{}, we use a normalized
$\rho$-weighted Jensen--Shannon divergence between $p_I$ and $p_H$. With
$\rho=0.8$ and
$m_\rho(u)=(1-\rho)p_I(u)+\rho p_H(u)$, define
\begin{equation}
\begin{aligned}
\mathcal D_{\mathrm{IHD}}(u)
=\frac{1}{H_{\mathrm b}(\rho)}
\Big[&
(1-\rho)\operatorname{KL}
\!\left(p_I(u)\,\|\,m_\rho(u)\right)\\
&+\rho\operatorname{KL}
\!\left(p_H(u)\,\|\,m_\rho(u)\right)
\Big].
\end{aligned}
\label{eq:ihd}
\end{equation}
Here
$H_{\mathrm b}(\rho)
=-(1-\rho)\log(1-\rho)-\rho\log\rho$
is the binary entropy. Because the weighted Jensen--Shannon divergence is
bounded by $H_{\mathrm b}(\rho)$,
$\mathcal D_{\mathrm{IHD}}(u)\in[0,1]$, with $\mathcal D_{\mathrm{IHD}}(u)=0$ if and only if
$p_I(u)=p_H(u)$.

Let
$\mathbf g(u)=
(\mathcal U_{\mathrm{MSP}}(u),
 \mathcal U_{\mathrm{loc}}(u),
 \mathcal D_{\mathrm{IHD}}(u))^\top$.
Using componentwise mean
$\boldsymbol\mu_{\mathrm{tr}}$, standard deviation
$\boldsymbol\sigma_{\mathrm{tr}}$ estimated once from predictions on the
corresponding source training set, and fixed weights
$\mathbf w=(1,0.5,0.2)^\top$, the final readout is
\begin{equation}
\mathcal U_{\mathrm{C\text{-}IHD}}(u)
=
\mathbf w^\top
\left[
\bigl(\mathbf g(u)-\boldsymbol\mu_{\mathrm{tr}}\bigr)
\oslash\boldsymbol\sigma_{\mathrm{tr}}
\right],
\label{eq:cihd}
\end{equation}
where $\oslash$ denotes componentwise division. All statistics and weights are
fixed before evaluation; source-frozen transfer reuses the source statistics
unchanged. \cihd{} leaves $p_H$ and the predicted segmentation unchanged and
requires neither a trainable error predictor nor an additional model forward
pass.

\section{Experiments}

\begin{figure*}[t]
\centering
\includegraphics[width=\textwidth]{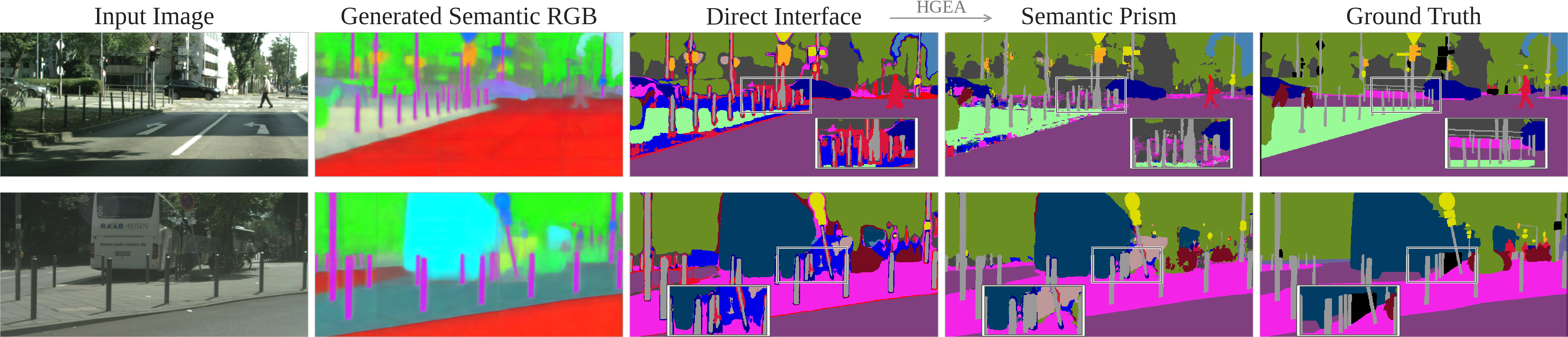}
\caption{Observable-to-final Cityscapes predictions. Fixed-codebook decoding
converts the generated semantic RGB image into the Direct Interface prediction;
\hgea{} then produces the final Semantic Prism prediction. Aligned
insets zoom in and highlight representative differences at semantic boundaries and thin
structures.}
\label{fig:observable-generation}
\end{figure*}

\subsection{Experimental setup}

Cityscapes \citep{cordts2016cityscapes} serves as our primary benchmark. Using
AdamW, we adapt the generator on its 2,975 training images for 100k steps
(learning rate $10^{-5}$, batch size one), then freeze it and train \hgea{} for
48k steps (learning rate $5\!\times\!10^{-4}$, weight decay $10^{-4}$,
gradient clipping 1.0). The training endpoints, interface temperature, and
\cihd{} coefficients are fixed without validation-set model selection, and
readout statistics are estimated from the corresponding source-training
predictions.

We evaluate the Cityscapes-trained model on all 500 validation images across
19 classes. We also train \method{} independently on BDD100K \citep{yu2020bdd100k} and evaluate the
full 1,000-image validation set. For source-frozen
transfer, the Cityscapes-trained model and fixed readout are applied unchanged
to all 406 ACDC \citep{sakaridis2021acdc} validation images, with ACDC annotations used only for
evaluation.

All evaluations use $1024\!\times\!512$ outputs, common label mappings, valid-pixel masks, and the same metric implementation, without multiscale or flip test-time augmentation or postprocessing. The comparison fixes output resolution, with method-specific inference costs reported in Table~\ref{tab:efficiency}.

We report mIoU, thin/rare-class mIoU, boundary F-score at a three-pixel
tolerance (BF@3), 15-bin expected calibration error (ECE$_{15}$), Brier score,
and negative log-likelihood (NLL). Pixel-error localization is evaluated using
the area under the receiver operating characteristic curve (AUROC), AUPR, and
the area under the risk--coverage curve (AURC), with incorrect valid pixels
treated as positives. External baselines use MSP for cross-model comparison;
the contribution of \cihd{} is evaluated against MSP on identical $p_H$.

\subsection{Observable semantic interface: evaluation on Cityscapes}

Before hierarchical refinement, fixed-codebook decoding of the generated semantic image attains 60.68\% mIoU, demonstrating that the visible output is independently evaluable. Figure~\ref{fig:observable-generation} shows that the rendered image captures major scene regions and illustrates representative boundary and thin-structure corrections by \hgea{}. The interface has 5.69\% ECE.

\begin{table*}[t]
\caption{Cityscapes val500 at $1024\!\times\!512$. Pixel-error scores are MSP
for external models and \cihd{} for \method{}. Ent.-AUPR and Mar.-AUPR use entropy and
inverse top-two margin, respectively. Baselines are SegFormer
\citep{xie2021segformer}, Mask2Former \citep{cheng2022mask2former}, DDPS
\citep{lai2023ddps}, DDP \citep{ji2023ddp}, and GSS \citep{chen2023gss}.}
\label{tab:main}
\centering
{\footnotesize
\setlength{\tabcolsep}{1.25pt}
\begin{tabular}{l l r r r r r r r r}
\toprule
& & \multicolumn{3}{c}{Segmentation} & \multicolumn{1}{c}{Calibration}
& \multicolumn{4}{c}{Pixel-Error Localization} \\
\cmidrule(lr){3-5}\cmidrule(lr){6-6}\cmidrule(lr){7-10}
Method & Paradigm & mIoU (\%)$\uparrow$ & Thin/Rare (\%)$\uparrow$ & BF@3 (\%)$\uparrow$
& ECE$_{15}$ (\%)$\downarrow$ & AUROC$\uparrow$ & AUPR$\uparrow$
& Ent.-AUPR$\uparrow$ & Mar.-AUPR$\uparrow$ \\
\midrule
SegFormer-B0 & Disc. & 71.13 & 61.90 & 79.51 & 1.08
& 0.9340 & 0.4419 & 0.4274 & 0.4225 \\
M2F-SwinT & Disc. & 77.92 & 71.03 & 85.63 & 1.80
& \textbf{0.9459} & 0.4339 & 0.4480 & 0.4163 \\
DDPS-B0 & Iter. Diff. & 74.17 & 67.06 & 82.83 & 1.48
& 0.9375 & 0.4291 & 0.4233 & 0.4093 \\
DDP-CNXT-T & Iter. Diff. & \textbf{79.10} & \textbf{73.33}
& \textbf{86.87} & 1.51 & 0.9442 & 0.4115 & 0.4045 & 0.3965 \\
GSS-FF\_R101 & Gen. Mask & 75.00 & 68.27 & 85.33 & -- & -- & -- & -- & -- \\
Direct Interface & 1-Step Gen. & 60.68 & 48.76 & 78.70 & 5.69 & -- & -- & -- & -- \\
\textbf{\method} & \textbf{1-Step Gen.} & 72.07 & 63.80 & 81.26
& \textbf{0.41} & 0.9457 & \textbf{0.4812}
& \textbf{0.4504} & \textbf{0.4546} \\
\bottomrule
\end{tabular}
}
\end{table*}

Table~\ref{tab:main} compares all methods under the common evaluation protocol.
Relative to direct-interface decoding, \hgea{} raises mIoU by 11.39 percentage points to
72.07\%, improves thin/rare-class mIoU and BF@3, and lowers ECE from 5.69\% to
0.41\%. Under the common-resolution protocol, \method{} achieves higher mIoU than SegFormer-B0 but remains below the stronger discriminative and diffusion-based baselines.

For pixel-error ranking, M2F-SwinT attains marginally higher AUROC, whereas \method{} with \cihd{} achieves the highest AUPR among the compared methods. Cross-model rankings reflect both predictor error distributions and uncertainty readouts; Table~\ref{tab:reliability-ablation} compares \cihd{} with MSP on identical $p_H$.

Additional codebook comparison, Brier score and NLL, temperature sensitivity, reliability diagrams, and spatial calibration analysis are provided in the supplementary material. Table~\ref{tab:matched-hierarchy} compares capacity-matched refiners to assess the contribution of multi-level generator features.

\begin{table*}[t]
\caption{Unified evaluation on BDD100K val1000. The MSP and \cihd{} rows share
$p_H$ and differ only in the uncertainty readout. T/R denotes thin/rare-class
mIoU. Bold and underlined values denote the best and second-best results.}
\label{tab:bdd100k}
\centering
{\footnotesize
\setlength{\tabcolsep}{3.0pt}
\begin{tabular}{@{}l l r r r r r r r@{}}
\toprule
& & \multicolumn{3}{c}{Segmentation} &
\multicolumn{2}{c}{Calibration} &
\multicolumn{2}{c}{Pixel-Error Localization} \\
\cmidrule(lr){3-5}\cmidrule(lr){6-7}\cmidrule(lr){8-9}
Method & Paradigm & mIoU (\%)$\uparrow$ & T/R (\%)$\uparrow$ & BF@3 (\%)$\uparrow$ & ECE$_{15}$ (\%)$\downarrow$ & Brier$\downarrow$ &
AUROC$\uparrow$ & AUPR$\uparrow$ \\
\midrule
DSNet-Base \citep{guo2024dsnet} & Disc. & \textbf{62.36} & \textbf{50.69} &
\textbf{74.08} & 3.70 & \underline{0.0985} & 0.8435 & 0.3776 \\
MSeg BDD-1M \citep{lambert2020mseg} & Disc. & 60.79 & 49.12 & 67.29 &
\underline{0.89} & 0.1003 & 0.8942 & 0.4299 \\
\textbf{\method{} (MSP)} & \textbf{1-Step Gen.} & \underline{62.22} &
\underline{50.19} & \underline{72.33} &
\textbf{0.88} & \textbf{0.0910} & \underline{0.9000} & \underline{0.4395} \\
\textbf{\method{} (\cihd{})} & \textbf{1-Step Gen.} & \underline{62.22} &
\underline{50.19} & \underline{72.33} &
\textbf{0.88} & \textbf{0.0910} & \textbf{0.9018} & \textbf{0.4481} \\
\bottomrule
\end{tabular}
}
\end{table*}

\subsection{In-domain evaluation on BDD100K}

For in-domain evaluation beyond Cityscapes, we train \method{} independently on
BDD100K and evaluate the full 1,000-image validation set \citep{yu2020bdd100k}.
As shown in Table~\ref{tab:bdd100k}, the model ranks second in mIoU,
thin/rare-class mIoU, and BF@3, trailing DSNet-Base by 0.14 and 1.75 percentage points in mIoU and BF@3, respectively, while achieving the lowest ECE and Brier score among
compared methods. On the same fixed $p_H$, \cihd{} raises AUROC/AUPR from
0.9000/0.4395 with MSP to 0.9018/0.4481 without changing segmentation or
calibration.

\subsection{Controlled attribution of hierarchical evidence}

To assess the contribution of generator features, we compare four capacity-matched refiners against the fixed Direct Interface (DI) baseline (Table~\ref{tab:matched-hierarchy}). Each refiner has
approximately 0.191M trainable parameters, with counts differing by less than
0.1\%, and is trained for 36k steps over three prespecified seeds using matched
sample order, crops, optimization, and losses.

The Observable-Interface Refiner (OI-Ref) receives only the rendered semantic
image and interface distribution $p_I$. The Capacity-Matched Flat Refiner
(CM-Flat) additionally receives the input RGB image and uses the same
interface-logit residual formulation, but no generator features. Single-Level
\hgea{} (SL-\hgea{}$_{\mathrm{mid}}$) uses only the prespecified middle
generator level, whereas Multi-Level \hgea{} (ML-\hgea{}) uses all three
levels.

\begin{table}[t]
\caption{Matched refinement ablation on Cityscapes val500. Learned controls
report three-seed mean $\pm$ sample standard deviation (both in \%). These checkpoints are trained separately from the canonical 72.07\%
result in Table~\ref{tab:main}.}
\label{tab:matched-hierarchy}
\centering
{\footnotesize
\setlength{\tabcolsep}{4.4pt}
\begin{tabular}{@{}l c c c c@{}}
\toprule
Model & mIoU$\uparrow$ & T/R$\uparrow$ & BF@3$\uparrow$
& ECE$_{15}\downarrow$ \\
\midrule
DI & 60.68 & 48.76 & 78.70 & 5.69 \\
\midrule
OI-Ref & \shortstack{67.58\\[-1pt]{\scriptsize$\pm0.88$}}
& \shortstack{57.79\\[-1pt]{\scriptsize$\pm1.19$}}
& \shortstack{79.65\\[-1pt]{\scriptsize$\pm0.36$}}
& \shortstack{0.40\\[-1pt]{\scriptsize$\pm0.02$}} \\
CM-Flat & \shortstack{69.75\\[-1pt]{\scriptsize$\pm0.40$}}
& \shortstack{60.83\\[-1pt]{\scriptsize$\pm0.63$}}
& \shortstack{79.82\\[-1pt]{\scriptsize$\pm0.23$}}
& \shortstack{0.47\\[-1pt]{\scriptsize$\pm0.14$}} \\
SL-\hgea{}$_{\mathrm{mid}}$
& \shortstack{70.54\\[-1pt]{\scriptsize$\pm0.59$}}
& \shortstack{61.77\\[-1pt]{\scriptsize$\pm0.75$}}
& \shortstack{\textbf{80.32}\\[-1pt]{\scriptsize$\pm0.56$}}
& \shortstack{\textbf{0.39}\\[-1pt]{\scriptsize$\pm0.12$}} \\
\textbf{ML-\hgea{}}
& \shortstack{\textbf{71.43}\\[-1pt]{\scriptsize$\pm0.47$}}
& \shortstack{\textbf{62.55}\\[-1pt]{\scriptsize$\pm1.17$}}
& \shortstack{80.26\\[-1pt]{\scriptsize$\pm1.25$}}
& \shortstack{0.43\\[-1pt]{\scriptsize$\pm0.16$}} \\
\bottomrule
\end{tabular}
}
\end{table}

Across all three seeds, mIoU follows the same ordering:
OI-Ref $<$ CM-Flat $<$ SL-\hgea{}$_{\mathrm{mid}}$ $<$ ML-\hgea{}.
ML-\hgea{} achieves $71.43\pm0.47$\% mIoU and outperforms CM-Flat and
SL-\hgea{}$_{\mathrm{mid}}$ by paired margins of $1.68\pm0.11$ and
$0.89\pm0.23$ mIoU points. Here, $\pm$ denotes the sample standard deviation
over the three seeds, computed from paired per-seed differences for the
reported margins. These mIoU margins quantify the additional gains of multi-level refinement over the capacity-matched flat and single-level alternatives. ML-\hgea{} also improves thin/rare-class mIoU and BF@3 over
CM-Flat while maintaining comparable ECE.

A pixel-wise comparison further shows that \hgea{} preserves the interface top-1
prediction on 93.48\% of valid pixels. Among interface errors, it
corrects 48.68\%, yielding a net pixel-accuracy gain of 3.67 percentage points. This gain
is concentrated at semantic boundaries, reaching 18.85 percentage points, compared with
1.67 percentage points in region interiors. Additional calibration, output-parameterization, removal, and
spatial-shuffling analyses are provided in the supplementary material.

\begin{figure*}[t]
\centering
\includegraphics[width=\textwidth]{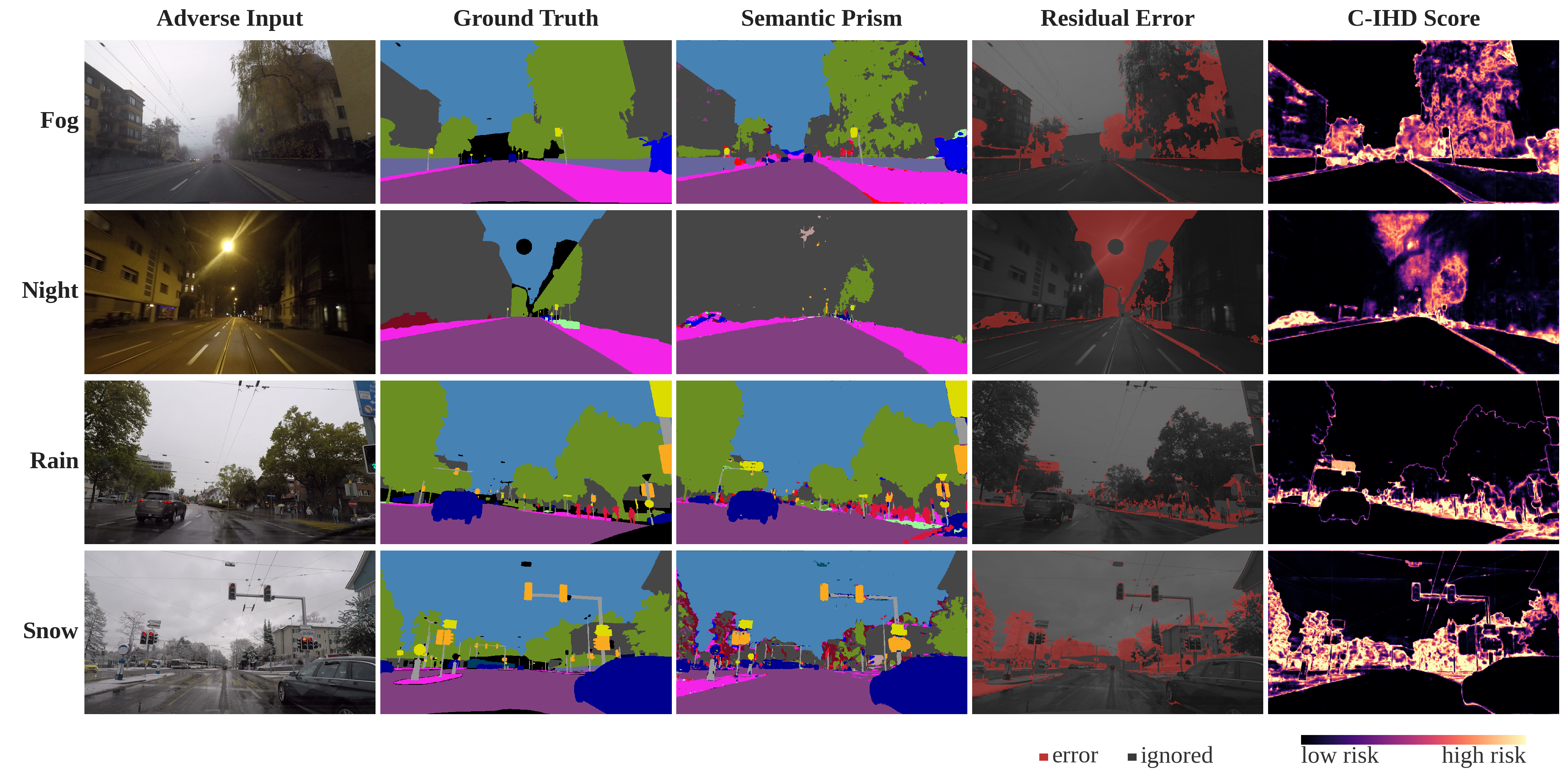}
\caption{Source-frozen ACDC transfer evaluation. Rows show scenarios of fog,
night, rain, and snow; columns show the input, ground truth, prediction,
pixel-error mask, and shared-scale \cihd{} risk map. Quantitative results are
reported in Table~\ref{tab:acdc}.}
\label{fig:acdc-qualitative}
\end{figure*}

\subsection{Source-frozen evaluation on ACDC}

We apply the Cityscapes-trained predictor and fixed \cihd{} readout unchanged to all 406 ACDC validation images spanning fog, night, rain, and snow, without target adaptation \citep{sakaridis2021acdc}. As shown in Table~\ref{tab:acdc}, \method{} attains 46.89\% mIoU and the lowest ECE in the comparison (8.48\%), while DDP-CNXT-T attains the highest mIoU. Cross-model rankings reflect different predictor error distributions; the paired \method{} rows compare the readouts on identical predictions. With $p_H$ held fixed, \cihd{} improves AUROC from 0.8903 to 0.9076 and AUPR from 0.6580 to 0.7557.

Figure~\ref{fig:acdc-qualitative} illustrates residual errors and \cihd{} risk near low-contrast boundaries, thin structures, and weather- or illumination-degraded regions. Examples include correctly localized errors, false positives along semantic boundaries, and confident errors that receive low risk.

\begin{table}[t]
\caption{Source-frozen ACDC val406 at $1024\!\times\!512$. The MSP and
\cihd{} rows share $p_H$ and differ only in the uncertainty readout. mIoU/ECE
are in \%. ``--'' denotes unavailable metrics. Baseline families follow
Table~\ref{tab:main}; the additional baselines are DANet\_R101\citep{fu2019danet} and DeepLabv3+\_R101\citep{chen2018deeplabv3plus}}
\label{tab:acdc}
\centering
{\footnotesize
\setlength{\tabcolsep}{1.3pt}
\begin{tabular}{@{}l c r r r r@{}}
\toprule
Method & Type & mIoU$\uparrow$ & ECE$_{15}\downarrow$ & AUROC$\uparrow$ & AUPR$\uparrow$ \\
\midrule
SegFormer-B0 & Disc. & 46.48 & 10.61 & 0.8407 & 0.5211 \\
DDPS-B0 & Diff. & 48.36 & 12.12 & 0.8164 & 0.4729 \\
M2F-SwinT & Disc. & 49.74 & 9.65 & 0.8965 & 0.5935 \\
GSS-FF\_R101 & Gen. & 37.26 & -- & -- & -- \\
DDP-CNXT-T & Diff. & \textbf{51.24} & 12.62 & 0.7109 & 0.3558 \\
DANet\_R101 & Disc. & 34.72 & 9.92 & 0.8829 & 0.6932 \\
DeepLabv3+\_R101 & Disc. & 39.67 & 10.22 & 0.8611 & 0.6080 \\
\textbf{Ours (MSP)} & \textbf{Gen.} & 46.89 & \textbf{8.48} &
0.8903 & 0.6580 \\
\textbf{Ours (\cihd{})} & \textbf{Gen.} & 46.89 & \textbf{8.48} &
\textbf{0.9076} & \textbf{0.7557} \\
\bottomrule
\end{tabular}
}
\end{table}

\subsection{Reliability at fixed predictions}

Across all three datasets, \cihd{} improves MSP-based error ranking without changing $p_H$. On Cityscapes val500, it increases AUROC from 0.94502 to 0.94572 and AUPR from 0.47814 to 0.48121, while reducing AURC from $4.281\times10^{-3}$ to $4.235\times10^{-3}$. Testing on BDD100K and ACDC shows the same
direction under in-domain training and source-frozen transfer, respectively.
Table~\ref{tab:reliability-ablation} shows that local context provides the main
individual gain; IHD lowers AUPR when combined with MSP alone, while its inclusion with local context improves AUROC and AUPR. Image-level paired bootstrap 95\% confidence intervals exclude
zero for the Cityscapes AUROC and AUPR gains (10,000 resamples). Model performance remains stable across the tested parameter ranges around
the training-set defaults. The supplement reports the
intervals, parameter ranges, standardization statistics, and selective-risk
operating points.

\begin{table}[t]
\caption{Cityscapes val500 uncertainty readouts at fixed $p_H$. MSP is
the no-\cihd{} baseline; AURC is in units of $10^{-3}$.}
\label{tab:reliability-ablation}
\centering
{\footnotesize
\setlength{\tabcolsep}{5pt}
\begin{tabular}{l r r r}
\toprule
Readout & AUROC$\uparrow$ & AUPR$\uparrow$ & AURC$\downarrow$ \\
\midrule
MSP & 0.94502 & 0.47814 & 4.281 \\
MSP + IHD & 0.94512 & 0.47730 & 4.273 \\
MSP + Local-MSP & 0.94558 & 0.48026 & 4.244 \\
\textbf{C-IHD} & \textbf{0.94572} & \textbf{0.48121} & \textbf{4.235} \\
\bottomrule
\end{tabular}
}
\end{table}

\subsection{Operational cost}

\begin{table}[t]
\caption{Operational cost on an A100 80GB (batch one, FP32/TF32; three
$512\!\times\!512$ crops). Components are timed independently against the
635-ms full path and are not additive.}
\label{tab:efficiency}
\centering
{\footnotesize
\setlength{\tabcolsep}{3.0pt}
\begin{tabular}{@{}l r r r@{}}
\toprule
\multicolumn{4}{@{}l}{\textbf{(a) Three-window throughput}} \\
Method & Steps/crop & FPS$\uparrow$ & GiB$\downarrow$ \\
\midrule
DDP-CNXT-T & 3 & \textbf{3.33} & \textbf{0.47} \\
DDPS-B0 & 20 & 0.84 & 1.25 \\
\textbf{\method} & \textbf{1} & 1.57 & 6.40 \\
\midrule
\multicolumn{4}{@{}l}{\textbf{(b) Incremental adaptation cost}} \\
Addition & Opt. params & Core (\%) & $\Delta$ms (\% full) \\
\midrule
LoRA (unmerged) & 9.002M & 0.698 & 128.93 (20.30) \\
VAE skips & 0.492M & 0.038 & 1.85 (0.29) \\
\hgea{} & 0.191M & 0.015 & 29.54 (4.65) \\
\cihd{} & 0 & 0 & 1.16 (0.18) \\
All optimized & 9.696M & 0.752 & -- \\
\bottomrule
\end{tabular}
}
\end{table}

Under the three-window protocol in Table~\ref{tab:efficiency}, the one-step pipeline achieves $1.88\times$ the throughput of 20-step DDPS; three-step DDP-CNXT-T has higher throughput and lower memory use. Training optimizes 9.696M parameters (0.752\% of the
core), including 0.012M existing conv-in weights. Unmerged LoRA is the largest
measured adaptation overhead; \hgea{} and \cihd{} account for 4.65\% and 0.18\%
of full-path latency, respectively. Full distributions, additional models, and
the LoRA merge audit are provided in the supplement.

\FloatBarrier
\section{Discussion and conclusion}

\method{} combines an independently evaluable semantic-image interface with feature-based refinement in the same logit space. A fixed codebook defines the interface distribution, and \hgea{} predicts additive logit corrections using aligned multi-level features together with image and interface inputs. On Cityscapes, the full refinement module yields an 11.39-point mIoU gain over direct-interface decoding; matched-capacity three-seed controls show higher mIoU for multi-level refinement than for flat and single-level alternatives. Evaluation with a separately trained BDD100K model and source-frozen ACDC transfer covers additional driving-data settings.

\cihd{} combines pointwise uncertainty, local context, and interface--refinement disagreement to improve AUPR over MSP on unchanged predictions across all three datasets, without an auxiliary predictor or additional forward pass. The interface provides the pre-refinement probability distribution, while the learned \hgea{} residual updates the corresponding class logits to produce the refined distribution. The current design relies on a closed-set codebook and a computationally demanding generator that requires multi-window inference; evaluation covers three driving datasets. The resulting pipeline exposes both interface and refined distributions for segmentation and post-hoc pixel-error ranking.

\FloatBarrier
\label{sec:main-end}

%% file: llm_usage_statement.tex
\section*{LLM Usage Statement}
OpenAI Codex (ChatGPT) was used for language polishing and grammar checking across the \textit{Abstract}, \textit{Introduction}, \textit{Related Work}, \textit{Method}, \textit{Experiments}, and \textit{Discussion and Conclusion} sections. It also assisted with parts of the code implementation and preliminary visualizations of the results. The authors reviewed all AI-assisted content and take full responsibility for the manuscript, code, and reported results.

%% file: supporting_materials.tex
This supplement details the implementation, evaluation protocols, and additional
analyses for \method{}. Unless stated otherwise, the Cityscapes analyses use
the same frozen generator, fixed semantic-RGB interface, and final
Hierarchical Generator Evidence Alignment (\hgea{}) checkpoint. We denote the Direct Interface and refined class distributions
by $p_I$ and $p_H$, respectively.

\section{Implementation and Evaluation Protocol}

\subsection{Software Environment and Reproducibility}

All experiments use Python 3.10, PyTorch 2.5.1~\citep{paszke2019pytorch}, TorchVision 0.20.1, and
CUDA 12.1 on a single NVIDIA GPU. The one-step generator uses SD-Turbo~\citep{sauer2023add} with
the pix2pix-Turbo adapter stack~\citep{parmar2024onestep}, implemented with
diffusers~0.38.0~\citep{vonplaten2022diffusers},
transformers~5.12.1~\citep{wolf2020transformers},
peft~0.19.1~\citep{mangrulkar2022peft},
accelerate~1.14.0~\citep{gugger2022accelerate}, and safetensors~0.8.0.
Other dependencies are NumPy~2.2.5~\citep{harris2020numpy}, Pillow~12.1.0,
scikit-image~0.25.2~\citep{vanderwalt2014skimage},
scikit-learn~1.7.2~\citep{pedregosa2011sklearn},
SciPy~1.15.3~\citep{virtanen2020scipy},
OpenCV~4.13~\citep{bradski2000opencv}, and
cityscapesscripts~2.2.4. The complete pinned specification is provided in a
Conda environment file. Training requires CUDA and approximately 24\,GB of
device memory for $512\!\times\!512$ crops. CPU-only execution is not supported.

Training has two stages: generator adaptation followed by \hgea{}
refinement, with the generator frozen throughout the second stage.
Table~\ref{tab:supp-training-config} summarizes the training settings
for both stages.

\begin{table}[t]
\caption{Two-stage training configuration for generator adaptation and HGEA refinement.}
\label{tab:supp-training-config}
\centering
{\small
\setlength{\tabcolsep}{4pt}
\begin{tabular}{l c c}
\toprule
\textbf{Configuration} & \textbf{Generator Adaptation} & \textbf{HGEA} \\
\midrule
Total steps & 100,000 & 48,000\\
Optimizer & AdamW & AdamW \\
Learning rate & $10^{-5}$ & $5\times10^{-4}$\\
Weight decay & $10^{-2}$ & $10^{-4}$ \\
Batch size & 1 & 1 \\
Precision & BF16 autocast & FP32 \\
Gradient clipping & 1.0 & 1.0 \\
Random seed & 13013 & 38001 \\
\bottomrule
\end{tabular}
}

\end{table}

We use fixed random seeds of 13013 for generator adaptation and 38001 for
\hgea{} training. Before constructing the model or data loader, we seed
Python and NumPy, along with PyTorch's CPU and CUDA random number
generators. During generator adaptation, we record the resolved
configuration and save checkpoints. Each \hgea{} checkpoint also stores
the random number generator states so they can be restored when training
resumes. If the loss becomes non-finite, generator adaptation stops.
\hgea{} training skips the affected optimizer update and continues from
the last valid state.

\subsection{Data Pipeline and Preprocessing}

We pair each Cityscapes~\citep{cordts2016cityscapes} \texttt{leftImg8bit}
image with its corresponding \texttt{gtFine\_labelTrainIds} annotation
following the \texttt{\{train,val\}/city/} directory structure. Images
are loaded as 8-bit RGB, and label maps use the 19-class training IDs,
with 255 marking ignored pixels. Spatial transformations are synchronized
for each image--label pair. Resizing uses antialiased bicubic interpolation
for images and nearest-neighbor interpolation for labels. The same crop
and horizontal flip are applied to both.

During generator adaptation, we sample a scale factor uniformly from
$[0.5,2.0]$. If the resized short side is less than 512 pixels, we upscale
the sample to accommodate a $512\!\times\!512$ crop. Generator targets
are constructed by mapping the cropped labels to the fixed semantic-RGB
codebook, with a reserved color for ignored pixels. The resulting RGB
values are scaled to $[-1,1]$. Rare-target sampling selects the window
with the greatest coverage of the designated thin or small classes from
a bounded set of candidates. If no candidate qualifies, we sample a crop uniformly. Evaluation uses
the fixed inference windows described in the next subsection.

\subsection{Full-Frame Inference and Stitching}

All reported outputs have a resolution of $1024\!\times\!512$.
\method{} processes three overlapping $512\!\times\!512$ windows
whose left edges are at horizontal coordinates of 0, 256, and 512 pixels.
For each window, we run one deterministic generator step, decode the
output with the fixed codebook, and apply \hgea{} refinement.
In overlapping regions, we average $p_I$ and $p_H$ separately across
windows and renormalize each distribution. Final segmentation labels
are obtained by taking the classwise argmax of the stitched $p_H$,
while \cihd{} is computed from the stitched $p_I$ and $p_H$.
Evaluation uses no multiscale or flip test-time augmentation,
additional postprocessing, or model ensembling.

\subsection{One-Step Generator Adaptation}

We initialize the semantic-image generator from pretrained
SD-Turbo~\citep{sauer2023add} with newly initialized pix2pix-Turbo
adapters~\citep{parmar2024onestep}. Adaptation updates rank-8 low-rank
adaptation (LoRA) weights~\citep{hu2022lora} in the
U-Net~\citep{ronneberger2015unet}, the U-Net input convolution, rank-4
LoRA weights in the variational autoencoder
(VAE)~\citep{kingma2014vae,rombach2022ldm}, and four VAE skip
convolutions, for a total of 9,505,160 trainable parameters.

We train for 100,000 steps using AdamW~\citep{loshchilov2019adamw}
with a constant learning rate of $10^{-5}$, weight decay of $10^{-2}$,
$(\beta_1,\beta_2)=(0.9,0.999)$, and $\epsilon=10^{-8}$.
Training uses batch size one, gradient clipping at 1.0, and bfloat16
autocast.

Data augmentation consists of scale jitter in $[0.5,2.0]$,
$512\!\times\!512$ crops, and horizontal flips with probability 0.5.
The first 20,000 steps use random crops. At each subsequent step, we
attempt a rare-target crop with probability 0.5, without resetting
the optimizer state.

Let $\Omega$ denote the image pixel grid and $y(u)$ the ground-truth
class label at pixel $u$. We exclude pixels with the ignore label 255,
defining $\Omega_{\mathrm v}=\{u\in\Omega:y(u)\ne255\}$.
For generator supervision, generated colors and codebook prototypes
are represented in normalized $[0,1]$ RGB coordinates. At each valid
pixel, $\mathbf s(u)$ is the generated color and
$\mathbf s^\star(u)=\mathbf c_{y(u)}$ is the target color, where
$\mathbf c_k$ is the prototype for class $k$. Class indices range over
the $K=19$ semantic classes. Following the main paper, we define
\begin{equation}
d_k^{\mathrm G}(\mathbf v)
=\|\mathbf v-\mathbf c_k\|_2^2,
\end{equation}
where $\mathbf v\in[0,1]^3$ is an RGB color vector. The generator
objective is
\begin{equation}
\begin{aligned}
\mathcal L_G={}&0.5\mathcal L_{\mathrm{rgb}}
+3.0\mathcal L_{\mathrm{proto}}
+\mathcal L_{\mathrm{valid}}\\
&+1.5\mathcal L_{\mathrm{margin}}
+0.3\mathcal L_{\mathrm{boundary}}.
\end{aligned}
\label{eq:supp-generator-objective}
\end{equation}

The RGB loss $\mathcal L_{\mathrm{rgb}}$ is Smooth-L1 between
$\mathbf s(u)$ and $\mathbf s^\star(u)$ over valid pixels.
The prototype loss $\mathcal L_{\mathrm{proto}}$ is class-weighted
cross-entropy computed from logits
$-d_k^{\mathrm G}(\mathbf s(u))/\tau_g$, where $\tau_g=0.03$ is
the training temperature. The validity loss
$\mathcal L_{\mathrm{valid}}$ averages the squared distance to the
nearest prototype, $\min_k d_k^{\mathrm G}(\mathbf s(u))$.
The margin loss $\mathcal L_{\mathrm{margin}}$ averages the penalty
\[
\left[
m+d_{y(u)}^{\mathrm G}(\mathbf s(u))
-\min_{j\ne y(u)}d_j^{\mathrm G}(\mathbf s(u))
\right]_+,
\]
where $m=0.02$ is the margin in normalized squared-distance units,
$j$ indexes competing classes, and $[a]_+=\max(a,0)$.
The boundary loss $\mathcal L_{\mathrm{boundary}}$ applies the same
margin penalty to valid boundary pixels. We construct the boundary mask by marking both pixels in each
horizontally or vertically adjacent pair with different ground-truth
labels, then dilate the mask with a $5\!\times\!5$ square kernel.
Ignored pixels are excluded from all loss reductions.

For the prototype and margin losses, we compute class weights as the
inverse square root of class frequency and divide them by their mean.
We then multiply the weights of the designated rare classes by 1.5,
cap all weights at 8, and divide them by their new mean so that the
average weight across classes is one. After adaptation, the generator
remains frozen throughout \hgea{} training.

\subsection{Fixed Semantic-RGB Interface}

We construct the High-Separation codebook $\mathcal C_{\mathrm{HS}}$
by greedy max--min selection on a saturated RGB grid, followed by
class assignment to improve color separation between commonly confused
classes.
The codebook is fixed before training and remains
unchanged at inference. Table~\ref{tab:supp-codebook} lists all prototypes
in prediction-channel order.

\begin{table*}[!t]
\caption{RGB prototypes in the High-Separation codebook.
Values are 8-bit RGB triplets. Classes appear in prediction-channel
order, read from left to right within each row and then from top
to bottom.}
\label{tab:supp-codebook}
\centering
{\small
\setlength{\tabcolsep}{4pt}
\begin{tabular*}{0.98\textwidth}{@{\extracolsep{\fill}}l c l c l c l c@{}}
\toprule
Class & RGB prototype & Class & RGB prototype & Class & RGB prototype & Class & RGB prototype \\
\midrule
road & $(255,36,36)$ & sidewalk & $(36,180,180)$ & building & $(132,180,36)$ & wall & $(36,36,255)$ \\
fence & $(180,255,36)$ & pole & $(228,36,255)$ & traffic light & $(255,180,36)$ & traffic sign & $(36,132,255)$ \\
vegetation & $(36,255,36)$ & terrain & $(255,255,132)$ & sky & $(132,255,255)$ & person & $(228,132,132)$ \\
rider & $(132,36,180)$ & car & $(132,132,255)$ & truck & $(255,36,132)$ & bus & $(36,255,255)$ \\
train & $(255,132,255)$ & motorcycle & $(132,255,132)$ & bicycle & $(36,255,132)$ & & \\
\bottomrule
\end{tabular*}
}
\end{table*}

For interface decoding, we quantize the generated semantic RGB image
to 8-bit values and store the quantized image losslessly. Each raw
generator output component $t\in[-1,1]$ is mapped to an integer by
\begin{equation}
Q_8(t)=\operatorname{round}_{\mathrm{even}}
\left(\operatorname{clip}(127.5(t+1),0,255)\right),
\label{eq:supp-q8}
\end{equation}
where $\operatorname{clip}$ restricts values to $[0,255]$ and
$\operatorname{round}_{\mathrm{even}}$ rounds to the nearest integer,
with ties resolved to the even integer.

For inference window $w$, let $\mathbf t_w(u)\in[-1,1]^3$ denote the
raw RGB output at pixel $u$. Applying $Q_8$ componentwise gives the
stored semantic RGB image $\mathbf s_w=Q_8(\mathbf t_w)$.
Let $\mathbf c_k^{(8)}$ denote the 8-bit prototype for class $k$
listed in Table~\ref{tab:supp-codebook}. Using the squared-distance
notation of the main paper, the per-window interface distribution is
\begin{equation}
\begin{aligned}
d_{w,k}^{\mathrm I}(u)
&=\|\mathbf s_w(u)-\mathbf c_k^{(8)}\|_2^2,\\
p_{I,k}^{(w)}(u)
&=\frac{\exp[-d_{w,k}^{\mathrm I}(u)/\tau_I]}
{\sum_j\exp[-d_{w,j}^{\mathrm I}(u)/\tau_I]}.
\end{aligned}
\label{eq:supp-interface-render}
\end{equation}
Here $k$ and $j$ index the 19 semantic classes, and
$p_{I,k}^{(w)}(u)$ is the interface probability of class $k$ at pixel
$u$ in window $w$. The interface temperature is fixed at $\tau_I=900$
in squared 8-bit RGB units. It is distinct from the supervision
temperature $\tau_g$, which uses squared distances in normalized
$[0,1]$ RGB space.

Table~\ref{tab:supp-codebook-comparison} compares Direct Interface
decoding with the Cityscapes (CS) palette and the High-Separation
codebook at fixed training endpoints. Relative to the CS palette,
the High-Separation codebook improves mean intersection over union
(mIoU) by 0.82 percentage points and the boundary F-score with a
three-pixel matching tolerance (BF@3) by 10.81 percentage points.

\begin{table}[t]
\caption{Comparison of two semantic-RGB codebooks for Direct Interface
decoding on the 500-image Cityscapes validation set. $d_{\min}$ is the
minimum distance between distinct prototypes. Bnd. gap is the mean of
$d_2-d_1$ over boundary pixels, where $d_1$ and $d_2$ are the distances
from the generated color to its nearest and second-nearest prototypes,
respectively. All distances are computed in 8-bit RGB space using
the Euclidean ($\ell_2$) metric.}
\label{tab:supp-codebook-comparison}
\centering
{\small
\setlength{\tabcolsep}{2pt}
\begin{tabular*}{\columnwidth}{@{\extracolsep{\fill}}l c c c c@{}}
\toprule
Codebook & mIoU (\%)$\uparrow$ & BF@3 (\%)$\uparrow$ & $d_{\min}\uparrow$ & Bnd. gap$\uparrow$ \\
\midrule
CS palette & 59.86 & 67.89 & 20.00 & 7.39 \\
High-Sep. (ours) & \textbf{60.68} & \textbf{78.70} & \textbf{89.04} & \textbf{27.12} \\
\bottomrule
\end{tabular*}
}
\end{table}

\subsection{HGEA Architecture and Optimization}

\paragraph{Feature extraction and fusion.}
For each $512\!\times\!512$ crop, we extract three feature maps from
the frozen VAE encoder. From fine to coarse, the maps have 128, 256,
and 512 channels at spatial resolutions of $256\!\times\!256$,
$128\!\times\!128$, and $64\!\times\!64$, respectively. Each map
passes through a $1\!\times\!1$ convolution that projects it to
24 channels, followed by group normalization with eight
groups~\citep{wu2018groupnorm} and a SiLU
activation~\citep{elfwing2018silu}. We resize the projected feature maps to the crop resolution using
bilinear interpolation with the outer boundaries of the input and
output pixel grids aligned. Concatenating the 72 projected
channels with the input RGB image (3 channels), the generated semantic RGB image (3 channels),
and the 19 interface probability maps gives a 97-channel input to the residual mapper.

\paragraph{Residual refinement.}
The residual mapper possesses two $3\!\times\!3$ convolutional blocks,
each with 96 output channels, group normalization with eight groups,
and SiLU. A zero-initialized $1\!\times\!1$ convolution maps the
final hidden features to 19 class-logit corrections. The feature
projections and residual mapper together contain 190,891 trainable
parameters.

For window $w$, the fixed-codebook logits are given by
$z_{w,k}^{I}(u)=-d_{w,k}^{\mathrm I}(u)/\tau_I$ using the squared
RGB distances in Eq.~\eqref{eq:supp-interface-render}.
We add the predicted residual $\Delta\mathbf z_w^H(u)$ to these
logits and apply a softmax over classes to obtain
\begin{equation}
p_H^{(w)}(u)=\operatorname{softmax}\!\left(
\mathbf z_w^I(u)+\Delta\mathbf z_w^H(u)
\right).
\end{equation}
At initialization, the residual is zero at every pixel, so
$p_H^{(w)}(u)=p_I^{(w)}(u)$. At inference, we compute the refined
probabilities for each window and stitch them on the full-frame grid.

\paragraph{Training objectives.}
We supervise the refined distribution $p_H^{(w)}$ using the
ground-truth labels at valid pixels in window $w$. The objective is
\begin{equation}
\begin{aligned}
\mathcal L_H ={}&
\mathcal L_{\mathrm{seg}}
+\lambda_{\mathrm{bnd}}\mathcal L_{\mathrm{bnd}}\\
&+\lambda_{\mathrm{fuse}}\mathcal L_{\mathrm{fuse}}
+\lambda_{\mathrm{fp}}\mathcal L_{\mathrm{fp}},
\end{aligned}
\label{eq:supp-hgea-objective}
\end{equation}
where the coefficients depend on the training phase.
The segmentation term $\mathcal L_{\mathrm{seg}}$ starts with
standard cross-entropy and uses class-weighted cross-entropy in the
later phases. The boundary term $\mathcal L_{\mathrm{bnd}}$ applies
cross-entropy within a band around ground-truth class boundaries.

The fusion term $\mathcal L_{\mathrm{fuse}}$ applies to valid
interior pixels where $\max_k p_{I,k}^{(w)}(u)\geq 0.90$.
It uses the soft target
\begin{equation}
\widetilde p^{(w)}(u)
=0.25\,p_I^{(w)}(u)+0.75\,p_H^{(w)}(u).
\label{eq:supp-hgea-fusion-target}
\end{equation}
The interface contribution anchors the prediction to the fixed
codebook at these high-confidence pixels.

\paragraph{False-positive suppression.}
In the final training phase, $\mathcal L_{\mathrm{fp}}$ penalizes
motorcycle and pole probabilities at valid pixels whose ground-truth
labels belong to a predefined set of negative classes. The term
applies only to crops sampled as foreground hard negatives that
contain no ground-truth motorcycle pixels.

\paragraph{Optimization and sampling.}
We keep the generator frozen and train the feature projections and
residual mapper for 48,000 steps using
AdamW~\citep{loshchilov2019adamw}. We use a learning rate of
$5\!\times\!10^{-4}$, weight decay of $10^{-4}$, and gradient
clipping at 1.0. We sample crops uniformly at the start of training
and introduce target-aware sampling in later phases. We evaluate
the checkpoint at the prespecified training endpoint.

\subsection{Metric Definitions}

The thin/rare subset comprises wall, fence, pole, traffic light,
traffic sign, terrain, rider, truck, bus, train, motorcycle, and
bicycle. Ignored pixels are excluded from all evaluations.

Segmentation metrics use a $19\!\times\!20$ confusion matrix
accumulated across images. The additional prediction column records
out-of-range labels, which contribute false negatives at valid
ground-truth pixels. mIoU averages classwise IoU over classes with
nonzero union. BF@3 uses a three-pixel boundary-matching tolerance.
Regional boundary statistics use a ground-truth boundary band formed
by four-neighbor label transitions and disk dilation at the stated
radius; interior statistics use the valid pixels outside this band.
Codebook validity is summarized by the mean and 95th-percentile
nearest-prototype distances and the fraction within the specified
tolerance.

Top-label expected calibration error (ECE) uses equal-width confidence
bins and weights each bin's calibration gap by the fraction of pixels
in that bin~\citep{guo2017calibration}. Adaptive top-label ECE uses
deterministic equal-mass rank bins~\citep{nixon2019calibration}.
We compute both ECE variants, the Brier score~\citep{brier1950verification},
and negative log-likelihood (NLL)~\citep{gneiting2007proper} over all
valid pixels after normalizing the class probabilities at each pixel.

For pixel-error ranking, we treat misclassified valid pixels as
positives and correctly classified valid pixels as negatives.
We report the area under the receiver operating characteristic curve
(AUROC) and the area under the precision--recall curve
(AUPR)~\citep{davis2006relationship}, using average precision as the
AUPR measure. For each image, we sample valid pixels without replacement
using an image-specific seed and reuse the same subset across checkpoints
and readouts.

We construct the risk--coverage curve by sorting pixels in order of
increasing uncertainty and computing the error rate among the
least uncertain pixels retained at each coverage
level~\citep{geifman2017selective}. Coverage is the fraction
of evaluated pixels retained. We compute the area under the risk--coverage
curve (AURC) as the mean retained error rate over all nonempty prefixes
of this ordering. R@$c$ denotes the retained error rate at coverage $c$.

\section{Controlled Refinement Analysis}

\subsection{Matched Controls and Hierarchy Interventions}

Table~\ref{tab:supp-matched-hierarchy} compares fixed-codebook decoding
with learned refiners that use different inputs. The Direct Interface (DI)
decodes the rendered semantic RGB image $\mathbf{s}$ using the fixed
codebook. The Observable-Interface Refiner (OI-Ref) uses $\mathbf{s}$
and the interface distribution $p_I$. The Capacity-Matched Flat Refiner
(CM-Flat) adds the input RGB image $\mathbf{x}$ to these two inputs.
Each Single-Level \hgea{} (SL-\hgea{}) variant uses features from one
level of the generator's VAE encoder, while Multi-Level \hgea{}
(ML-\hgea{}) uses all three levels.

The learned refiners follow a common 36,000-step training protocol.
For each seed, they use the same sample order, crops, optimizer settings,
losses, and evaluation code. The trainable parameter counts of the six
learned refiners in the upper block differ by less than 0.5\%.

\begin{table*}[!t]
\caption{Capacity-matched refinement controls on Cityscapes val500.
The upper block reports results for seed 18313. The lower block gives
per-seed mIoU and the mean $\pm$ sample standard deviation across three
seeds. Levels 1--3 denote the fine, middle, and coarse VAE encoder
levels, respectively. Thin/rare mIoU is evaluated on the designated
class subset. BF@3 is the boundary F-score at a three-pixel matching
tolerance. The canonical checkpoint (72.07\% mIoU) is evaluated
separately.}
\label{tab:supp-matched-hierarchy}
\centering
{\small
\setlength{\tabcolsep}{3.2pt}
\textbf{Matched comparison (seed 18313)}\\[2pt]
\begin{tabular*}{0.98\textwidth}{@{\extracolsep{\fill}}l l c c c c c c c@{}}
\toprule
Model & Evidence & \shortstack{Trainable refiner\\parameters}
& mIoU (\%)$\uparrow$ & \shortstack{Thin/rare\\mIoU (\%)$\uparrow$}
& BF@3 (\%)$\uparrow$ & ECE (\%)$\downarrow$
& Brier$\downarrow$ & NLL$\downarrow$ \\
\midrule
DI & Fixed codebook & 0 & 60.68 & 48.76 & 78.70 & 5.69 & 0.144 & 1.143 \\
OI-Ref & $\mathbf{s},p_I$ & $\approx$190.9k & 67.70 & 57.84 & 79.71 & \textbf{0.41} & 0.083 & 0.207 \\
CM-Flat & $\mathbf{x},\mathbf{s},p_I$ & 190,915 & 70.19 & 61.50 & 80.09 & 0.51 & 0.081 & 0.200 \\
SL-\hgea{}$_{\mathrm{fine}}$ & Level 1 & 190,395 & 69.99 & 61.30 & 79.80 & 0.58 & 0.081 & 0.204 \\
SL-\hgea{}$_{\mathrm{mid}}$ & Level 2 & 190,763 & 71.14 & 62.56 & 80.89 & 0.45 & 0.077 & 0.185 \\
SL-\hgea{}$_{\mathrm{coarse}}$ & Level 3 & 190,843 & 66.76 & 56.70 & 78.43 & 0.84 & 0.089 & 0.257 \\
ML-\hgea{} & Levels 1--3 & 190,891 & \textbf{71.97} & \textbf{63.67} & \textbf{81.22} & 0.42 & \textbf{0.075} & \textbf{0.178} \\
\bottomrule
\end{tabular*}

\par\medskip
\textbf{Per-seed mIoU (\%)$\uparrow$}\\[2pt]
\setlength{\tabcolsep}{8pt}
\begin{tabular*}{0.98\textwidth}{@{\extracolsep{\fill}}l c c c c@{}}
\toprule
Seed & OI-Ref & CM-Flat & SL-\hgea{}$_{\mathrm{mid}}$ & ML-\hgea{} \\
\midrule
18313 & 67.70 & 70.19 & 71.14 & 71.97 \\
28313 & 68.39 & 69.66 & 70.54 & 71.23 \\
38313 & 66.65 & 69.41 & 69.95 & 71.09 \\
\midrule
Mean $\pm$ SD & $67.58 \pm 0.88$ & $69.75 \pm 0.40$
& $70.54 \pm 0.59$ & $\mathbf{71.43} \pm 0.47$ \\
\bottomrule
\end{tabular*}
\par}
\end{table*}

For seed 18313, the middle encoder level gives the highest mIoU among
the single-level variants (71.14\%). ML-\hgea{} reaches 71.97\%, a gain
of 0.83 percentage points. It also gives the highest thin/rare mIoU
and BF@3 and the lowest Brier score and NLL in this run. OI-Ref has
the lowest ECE. Among the four refiners evaluated across all three
seeds, ML-\hgea{} ranks first in mIoU at each seed.

We compare two ways of predicting class logits under the matched
36,000-step protocol (Table~\ref{tab:supp-output-head}). The two heads
use the same three VAE encoder levels, decoder architecture, and
trainable parameter count. The Unconstrained Head (UC-Head) predicts
the final class logits directly. ML-\hgea{} predicts an additive
correction to the fixed-codebook interface logits $\mathbf{z}^I$.

\begin{table}[t]
\caption{Matched output-head comparison on Cityscapes val500 for
seed 18313. Thin/rare denotes mIoU over the thin/rare class subset.
Both mIoU columns, BF@3, and ECE are reported as percentages.}
\label{tab:supp-output-head}
\centering
{\small
\setlength{\tabcolsep}{2.2pt}
\begin{tabular*}{\columnwidth}{@{\extracolsep{\fill}}l c c c c c c@{}}
\toprule
Model & mIoU$\uparrow$ & Thin/rare$\uparrow$ & BF@3$\uparrow$
& ECE$\downarrow$ & Brier$\downarrow$ & NLL$\downarrow$ \\
\midrule
ML-\hgea{} & 71.97 & 63.67 & 81.22 & 0.42 & 0.075 & 0.178 \\
UC-Head & \textbf{73.06} & \textbf{65.12} & \textbf{82.50}
& \textbf{0.34} & \textbf{0.071} & \textbf{0.159} \\
\bottomrule
\end{tabular*}
\par}
\end{table}

In this single-seed comparison, UC-Head improves mIoU by 1.09
percentage points. It also raises thin/rare mIoU and BF@3 and lowers
ECE, Brier score, and NLL.

To assess how the trained refiner uses hierarchical features, we
remove the features or shuffle their spatial correspondence at
inference (Table~\ref{tab:supp-hierarchy-intervention}).

\begin{table}[t]
\caption{Hierarchy removal and spatial shuffling at a fixed
checkpoint. All configurations are evaluated on the same 100
Cityscapes validation images without retraining. Thin/rare denotes
mIoU over the thin/rare class subset. Both mIoU columns and BF@3 are
reported as percentages.}
\label{tab:supp-hierarchy-intervention}
\centering
{\small
\setlength{\tabcolsep}{2.1pt}
\begin{tabular*}{\columnwidth}{@{\extracolsep{\fill}}l c c c c c@{}}
\toprule
Configuration & mIoU$\uparrow$ & Thin/rare$\uparrow$ & BF@3$\uparrow$
& Brier$\downarrow$ & NLL$\downarrow$ \\
\midrule
Aligned hierarchy & \textbf{72.98} & \textbf{65.35} & \textbf{82.57}
& \textbf{0.087} & \textbf{0.217} \\
Hierarchy removed & 64.84 & 54.52 & 78.40 & 0.123 & 0.390 \\
Spatially shuffled & 59.44 & 51.36 & 71.77 & 0.170 & 0.451 \\
\bottomrule
\end{tabular*}
\par}
\end{table}

Relative to the aligned hierarchy, removing the features lowers mIoU
by 8.14 percentage points, and shuffling their spatial correspondence
lowers it by 13.54 points. On this subset, the trained refiner is
sensitive to both the presence and spatial alignment of the
hierarchical features.

\subsection{Spatial and Classwise Effects of Refinement}

We compare Direct Interface predictions with those of the canonical
\hgea{} checkpoint on all 500 Cityscapes validation images
(Table~\ref{tab:supp-regional-anatomy}). The ground-truth boundary band
is formed by dilating four-neighbor label transitions with a disk of
radius three pixels. The remaining valid pixels form the interior.

\begin{table*}[!t]
\caption{Regional accuracy, calibration, and prototype distances on
Cityscapes val500. The Euclidean distances $d_1$ and $d_2$ are measured
from each generated color to its nearest and second-nearest prototypes,
respectively, using continuous RGB values on the 0--255 scale before
rounding. The gap is $d_2-d_1$. Accuracy and equal-width 15-bin ECE use
the same valid pixels. Corr. is the fraction of Direct Interface errors
corrected by \hgea{}. Regr. is the fraction of initially correct Direct
Interface predictions that become incorrect after refinement. Net is
the change in accuracy in percentage points (pp).}
\label{tab:supp-regional-anatomy}
\label{tab:supp-spatial-calibration}
\label{tab:supp-rgb-geometry}
\centering
{\small
\setlength{\tabcolsep}{3.0pt}
\begin{tabular*}{0.98\textwidth}{@{\extracolsep{\fill}}l r r r r r r r r r r@{}}
\toprule
& \multicolumn{3}{c}{Prototype distances}
& \multicolumn{2}{c}{Direct Interface $p_I$}
& \multicolumn{2}{c}{Refined $p_H$}
& \multicolumn{3}{c}{Pixel transitions} \\
\cmidrule(lr){2-4}\cmidrule(lr){5-6}\cmidrule(lr){7-8}\cmidrule(lr){9-11}
Region & Mean $d_1$ & Median $d_1$ & Mean gap
& Acc. (\%) & ECE (\%) & Acc. (\%) & ECE (\%)
& Corr. (\%) & Regr. (\%) & Net (pp) \\
\midrule
Global & -- & -- & -- & 91.40 & 5.69 & 95.07 & 0.41 & 48.68 & 0.57 & +3.67 \\
Boundary & 67.56 & 69.99 & 27.12 & 54.92 & 31.22 & 73.77 & 4.46 & 48.03 & 5.09 & +18.85 \\
Interior & 31.34 & 21.69 & 75.54 & 96.19 & 2.36 & 97.86 & 0.42 & 49.70 & 0.23 & +1.67 \\
\bottomrule
\end{tabular*}
\par}
\end{table*}

\hgea{} improves pixel accuracy by 18.85 percentage points in the
boundary band and by 1.67 points in the interior. It corrects nearly
half of the Direct Interface errors in each region. The larger
boundary gain reflects the higher initial error rate in the band.
Among pixels initially classified correctly, 5.09\% in the boundary
band and 0.23\% in the interior become incorrect after refinement.

The vegetation class has the highest correction rate: \hgea{} corrects
61.45\% of the Direct Interface errors on pixels labeled as vegetation.
The number of correctly classified pixels decreases after refinement
for the person, motorcycle, truck, and rider classes.

\section{Interface Confidence and Calibration}

\subsection{Temperature Sensitivity}

Within each inference window, a positive softmax temperature preserves
the nearest-prototype class. In overlapping regions, averaging the
resulting probabilities across windows can change the class selected
after stitching.

\begin{table}[t]
\caption{Direct Interface sensitivity to temperature on Cityscapes
val500. Bold marks the fixed operating point $\tau_I=900$.
Disagreement is the fraction of pixel predictions that differ from this
operating point after full-frame probability stitching.}
\label{tab:supp-temperature}
\centering
{\small
\setlength{\tabcolsep}{2.5pt}
\begin{tabular*}{\columnwidth}{@{\extracolsep{\fill}}c c c c c c@{}}
\toprule
$\tau_I$ & mIoU (\%)$\uparrow$ & Conf. (\%) & ECE (\%)$\downarrow$ & NLL$\downarrow$ & Disagr. (\%) \\
\midrule
225 & 60.702 & 98.576 & 7.170 & 1.787 & 0.139 \\
450 & 60.698 & 98.140 & 6.732 & 1.502 & 0.068 \\
675 & 60.690 & 97.640 & 6.235 & 1.300 & 0.030 \\
\textbf{900} & \textbf{60.681} & \textbf{97.089} & \textbf{5.691} & \textbf{1.143} & \textbf{0.000} \\
1200 & 60.669 & 96.267 & 4.877 & 0.981 & 0.035 \\
1800 & 60.650 & 94.153 & 2.780 & 0.766 & 0.089 \\
3600 & 60.602 & 82.565 & 9.039 & 0.591 & 0.185 \\
\bottomrule
\end{tabular*}
}

\end{table}

At the fixed checkpoint, changing $\tau_I$ moves mIoU by at most
0.079 percentage points relative to the default, and fewer than
0.19\% of pixel labels change (Table~\ref{tab:supp-temperature}).
BF@3 ranges from 78.674\% to 78.702\%. Probability estimates are more
sensitive. Brier scores range from 0.132 to 0.160, and ECE is lowest
at $\tau_I=1800$ among the temperatures tested.

\subsection{Spatial Calibration and RGB Geometry}

At the operating temperature $\tau_I=900$, the Direct Interface has
an expected calibration error (ECE) of 31.22\% in the ground-truth
boundary band and 2.36\% in the interior
(Table~\ref{tab:supp-regional-anatomy}). Near boundaries, generated
colors are farther from their nearest prototypes and the gap between
the nearest and second-nearest distances is smaller, indicating less
separation between the two closest prototypes. After \hgea{}
refinement, boundary ECE falls to 4.46\%.

\begin{figure*}[t]
\centering
\includegraphics[width=0.39\textwidth]{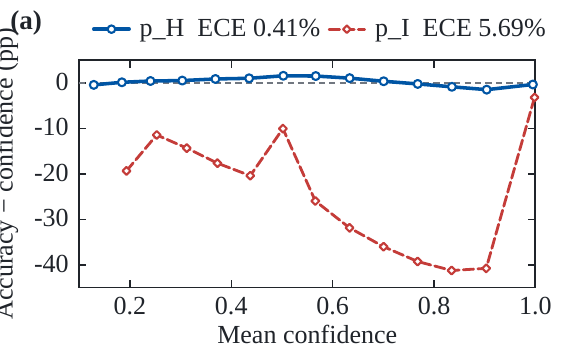}
\hfill
\includegraphics[width=0.57\textwidth]{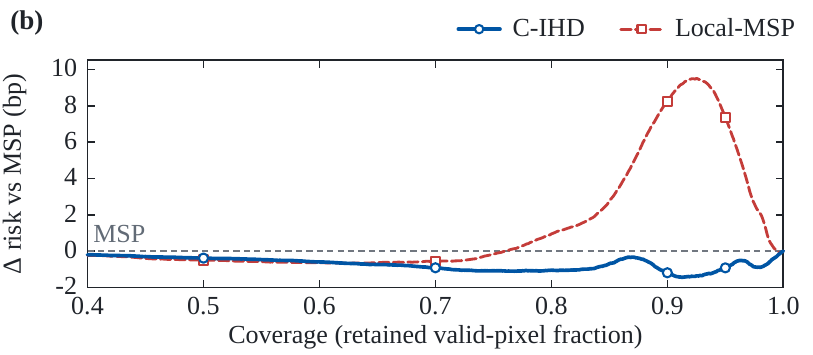}
\caption{Calibration and selective prediction on Cityscapes val500.
(a) Calibration residuals for the Direct Interface distribution $p_I$
and the canonical refined distribution $p_H$. Each residual is the
bin accuracy minus the mean confidence, in percentage points.
(b) Retained-risk differences for Local-MSP and \cihd{} relative to
MSP, in basis points, with $p_H$ held fixed. Negative values indicate
lower retained risk than MSP. Table~\ref{tab:supp-error-ranking}
reports AURC and retained risk at 50\%, 70\%, 90\%, and 95\% coverage.}
\label{fig:supp-reliability-selective}
\end{figure*}

Figure~\ref{fig:supp-reliability-selective}(a) compares calibration
residuals across confidence bins for the full validation set. On the
same refined predictions, equal-width top-label ECE ranges from
0.403\% to 0.407\%, and adaptive top-label ECE ranges from 0.383\%
to 0.405\% across the five bin counts tested (10, 15, 20, 30, and 50).
With 15 bins, adaptive top-label ECE is 0.405\%. The mean one-vs-rest
ECE is 0.122\%. This measure treats each class as a binary outcome
and averages the resulting ECE values across classes. Figure~\ref{fig:supp-reliability-selective}(b) shows that \cihd{}
reduces retained risk relative to MSP at all four reported coverage
levels. At 50\% coverage, Local-MSP gives a slightly larger reduction.

\section{Fixed-Prediction Reliability Analysis}

\subsection{C-IHD Definition and Component Analysis}

Contextual Interface Hierarchy Disagreement (\cihd{}) combines
pointwise uncertainty, its local average, and disagreement between
the interface and refined distributions. At pixel $u$, the maximum
softmax probability (MSP) readout uses its complement as the
uncertainty score~\citep{hendrycks2017baseline},
$\mathcal U_{\mathrm{MSP}}(u)=1-\max_k p_{H,k}(u)$.
Local-MSP, denoted by $\mathcal U_{\mathrm{loc}}(u)$, averages this
score over a $5\!\times\!5$ neighborhood.

Interface--hierarchy disagreement (IHD), denoted by
$\mathcal D_{\mathrm{IHD}}(u)$, is the weighted Jensen--Shannon
divergence~\citep{lin1991divergence} between $p_I(u)$ and $p_H(u)$,
with weights $1-\rho$ and $\rho$, respectively, divided by the binary
entropy $H_b(\rho)$. We set $\rho=0.8$. The unnormalized divergence
equals the mutual information between a Bernoulli selector and a class
drawn from the selected distribution. This mutual information is
bounded by $H_b(\rho)$, so $\mathcal D_{\mathrm{IHD}}(u)\in[0,1]$.

Let $\mathbf g(u)=(\mathcal U_{\mathrm{MSP}}(u),
\mathcal U_{\mathrm{loc}}(u),\mathcal D_{\mathrm{IHD}}(u))^\top$ and
$\mathbf w=(1,0.5,0.2)^\top$. The readout is
\begin{equation}
\mathcal U_{\mathrm{C\text{-}IHD}}(u)
=\mathbf w^\top
\bigl[(\mathbf g(u)-\boldsymbol\mu_{\mathrm{tr}})
\oslash\boldsymbol\sigma_{\mathrm{tr}}\bigr],
\label{eq:supp-cihd}
\end{equation}
where $\oslash$ denotes componentwise division. The componentwise
means and standard deviations estimated from Cityscapes training
predictions are
\begin{equation*}
\begin{aligned}
\boldsymbol\mu_{\mathrm{tr}}&=(0.041874,0.042058,0.051223)^\top,\\
\boldsymbol\sigma_{\mathrm{tr}}&=(0.115402,0.106081,0.173971)^\top.
\end{aligned}
\end{equation*}
We use these statistics and weights for Cityscapes validation and for
transfer of the frozen Cityscapes model to the Adverse Conditions Dataset with Correspondences
(ACDC)~\citep{sakaridis2021acdc}.

\begin{table*}[!t]
\caption{Pixel-error ranking on Cityscapes val500 with the canonical
refined distribution $p_H$ held fixed. R@50, R@70, R@90, and R@95
denote retained risk at 50\%, 70\%, 90\%, and 95\% coverage,
respectively. AURC and retained risk are reported in units of $10^{-3}$.}
\label{tab:supp-error-ranking}
\centering
{\small
\setlength{\tabcolsep}{4.7pt}
\begin{tabular*}{0.98\textwidth}{@{\extracolsep{\fill}}l c c c c c c c@{}}
\toprule
Readout & AUROC$\uparrow$ & AUPR$\uparrow$ & AURC$\downarrow$
& R@50$\downarrow$ & R@70$\downarrow$ & R@90$\downarrow$
& R@95$\downarrow$ \\
\midrule
MSP & 0.9450 & 0.4781 & 4.281 & 0.66 & 1.96 & 13.77 & 26.14 \\
Entropy & 0.9424 & 0.4504 & 4.417 & 0.66 & 1.98 & 14.28 & 27.98 \\
Margin & 0.9446 & 0.4546 & 4.301 & 0.65 & 1.95 & 13.75 & 26.31 \\
Local-MSP & 0.9434 & 0.4667 & 4.355 & \textbf{0.61} & 1.90 & 14.59 & 26.88 \\
IHD only & 0.9296 & 0.3014 & 5.080 & 0.63 & 2.02 & 17.37 & 34.11 \\
\cihd{} & \textbf{0.9457} & \textbf{0.4812} & \textbf{4.235}
& 0.62 & \textbf{1.87} & \textbf{13.65} & \textbf{26.05} \\
\bottomrule
\end{tabular*}
\par}
\end{table*}

Among the six readouts in Table~\ref{tab:supp-error-ranking},
\cihd{} demonstrates the highest AUROC and AUPR. Relative to MSP, it reduces
AURC from $4.281\!\times\!10^{-3}$ to $4.235\!\times\!10^{-3}$,
a decrease of about 1.1\%. It gives the lowest retained risk at 70\%,
90\%, and 95\% coverage. Local-MSP gives the lowest risk at 50\%
coverage. IHD alone shows the lowest AUROC and AUPR and the highest
AURC in this comparison.

A paired image-level bootstrap~\citep{efron1979bootstrap} with 10,000
resamples of the 500 Cityscapes validation images gives 95\% confidence
intervals of $[0.370,1.022]\!\times\!10^{-3}$ for the AUROC gain of
\cihd{} over MSP and $[2.085,4.056]\!\times\!10^{-3}$ for the AUPR
gain. These intervals quantify sampling variability across validation
images for the fixed checkpoint.

\subsection{Prespecified Readout Sensitivity Analysis}

We vary one setting at a time around the prespecified default.
We test $\rho\in\{0.20,0.50,0.65,0.80,0.90,0.95\}$,
Local-MSP weights in $\{0,0.25,0.50,0.75,1.00\}$, IHD weights in
$\{0,0.10,0.20,0.30,0.40\}$, and local-window side lengths in
$\{1,3,5,7,11\}$ pixels. We also evaluate the readout with
standardization disabled. Across the tested settings, AUROC ranges from 0.9451 to 0.9458,
AUPR from 0.4773 to 0.4814, and AURC from
$4.232\!\times\!10^{-3}$ to $4.273\!\times\!10^{-3}$.
The reported comparisons use the prespecified default.

\section{Cross-Domain Evaluation and Efficiency}

\subsection{ACDC Evaluation without Adaptation}

We apply the Cityscapes-trained model to all 406 ACDC validation
images without adaptation. The four readouts rank pixel errors using
the same refined distribution $p_H$ and the same segmentation labels
(Table~\ref{tab:supp-acdc-readout}).

\begin{table}[t]
\caption{Pixel-error ranking on ACDC val406 with the predictions of
the frozen Cityscapes model held fixed. AURC is reported in units
of $10^{-3}$.}
\label{tab:supp-acdc-readout}
\centering
{\small
\setlength{\tabcolsep}{4.2pt}
\begin{tabular}{@{}l c c c@{}}
\toprule
Readout & AUROC$\uparrow$ & AUPR$\uparrow$ & AURC$\downarrow$ \\
\midrule
MSP & 0.8903 & 0.6580 & 49.688 \\
MSP + IHD & 0.8910 & 0.6554 & 49.336 \\
MSP + Local-MSP & 0.8919 & 0.6657 & 49.282 \\
\cihd{} & \textbf{0.9076} & \textbf{0.7557} & \textbf{43.079} \\
\bottomrule
\end{tabular}
\par}
\end{table}

\cihd{} demonstrates the highest AUROC and AUPR among the four readouts and
lowers AURC by $6.609\!\times\!10^{-3}$ relative to MSP. Adding
IHD to MSP lowers AURC by $0.352\!\times\!10^{-3}$, and adding
Local-MSP lowers it by $0.406\!\times\!10^{-3}$.

\begin{table}[t]
\caption{Segmentation and \cihd{} pixel-error ranking by ACDC
condition. $N$ is the number of images, and Error is the percentage
of valid pixels classified incorrectly.}
\label{tab:supp-acdc-conditions}
\centering
{\small
\setlength{\tabcolsep}{2.4pt}
\begin{tabular}{@{}l c c c c c@{}}
\toprule
Condition & $N$ & mIoU (\%) & Error (\%) & AUROC & AUPR \\
\midrule
\textbf{Overall} & \textbf{406} & \textbf{46.89} & \textbf{19.69}
& \textbf{0.9076} & \textbf{0.7557} \\
\midrule
Fog & 100 & 58.15 & 11.85 & 0.9134 & 0.7017 \\
Night & 106 & 27.81 & 35.10 & 0.8521 & 0.7711 \\
Rain & 100 & 51.01 & 9.18 & 0.9186 & 0.6505 \\
Snow & 100 & 44.47 & 22.16 & 0.9076 & 0.8032 \\
\bottomrule
\end{tabular}
\par}
\end{table}

Among the four ACDC conditions, the \textit{night} subset has the highest
pixel error rate and the lowest mIoU and AUROC
(Table~\ref{tab:supp-acdc-conditions}). It ranks second in AUPR, after
the \textit{snow} subset. Since misclassified pixels are the positive examples,
differences in pixel error rates help put the AUPR values in
context~\citep{saito2015precision}.

\subsection{Inference and Adaptation Cost}

We measure inference latency at batch size one on a single NVIDIA
A100 PCIe GPU with 80\,GB of memory. Inference runs in FP32 with
TF32 enabled and autocast disabled. We average latency over 100
frames after 10 warm-up frames. The measurement includes
preprocessing, host-to-device transfer, model execution, and
stitching the window probabilities into a $1024\!\times\!512$ output.
Table~\ref{tab:supp-efficiency} reports inference throughput,
GPU memory use, optimized parameter counts, and independently
measured component latency differences.

\begin{table}[t]
\caption{Inference and adaptation costs under the three-window
protocol. Each latency difference $\Delta t$ in panel (b) is measured
independently. The percentages in parentheses are relative to the
full pipeline latency of 635.04\,ms. Core (\%) expresses parameter
counts as a percentage of the generator core. The optimized total is
computed before rounding and includes 0.012M existing U-Net
input-convolution parameters.}
\label{tab:supp-efficiency}
\centering
{\small
\setlength{\tabcolsep}{3pt}
\begin{tabular}{@{}l r r r@{}}
\toprule
\multicolumn{4}{@{}l}{\textbf{(a) Three-window inference}}\\
Method & Steps/crop & FPS$\uparrow$ & GPU mem. (GiB)$\downarrow$ \\
\midrule
DDP-CNXT-T & 3 & \textbf{3.33} & \textbf{0.47} \\
DDPS-B0 & 20 & 0.84 & 1.25 \\
\method{} & \textbf{1} & 1.57 & 6.40 \\
\midrule
\multicolumn{4}{@{}l}{\textbf{(b) Parameter counts and latency}}\\
Component & Opt. params & Core (\%) & $\Delta t$ (ms) \\
\midrule
LoRA (unmerged) & 9.002M & 0.698 & 128.93 (20.30\%) \\
VAE skips & 0.492M & 0.038 & 1.85 (0.29\%) \\
\hgea{} & 0.191M & 0.015 & 29.54 (4.65\%) \\
\cihd{} & 0 & 0 & 1.16 (0.18\%) \\
All optimized & 9.696M & 0.752 & -- \\
\bottomrule
\end{tabular}
\par}
\end{table}

Under the three-window protocol, \method{} runs at 1.57\,FPS,
about 1.9 times the throughput of 20-step
DDPS-B0~\citep{lai2023ddps}. Three-step
DDP-CNXT-T~\citep{ji2023ddp} reaches 3.33\,FPS and uses less GPU
memory than \method{} (Table~\ref{tab:supp-efficiency}).

Unmerged LoRA accounts for most of the optimized parameters and
produces the largest latency difference in the component comparisons.
The independently measured latency differences are 29.54\,ms for
\hgea{} and 1.16\,ms for \cihd{}, which has no trainable parameters.
We benchmark \method{} with unmerged LoRA because the tested merged
variant failed the prespecified check for numerical equivalence.

\section{Artifact Integrity and Metric Reproduction}

We checked the stored refined distributions $p_H$ and lossless semantic
RGB images for all 500 Cityscapes validation images against their
manifests. We then recomputed mIoU from these outputs over 228,972,480
valid pixels. Decoding the stored RGB images with the fixed codebook
at $\tau_I=900$ reproduced the Direct Interface result of 60.68\%.
Evaluating the labels obtained from the stored $p_H$ distributions
reproduced the refined result of 72.07\%.